\PassOptionsToPackage{pagebackref,breaklinks,colorlinks,allcolors=cvprblue}{hyperref}
\documentclass[runningheads]{llncs}

\usepackage[arxiv]{eccv}

\usepackage{colortbl}
\usepackage{multirow}
\usepackage{caption}
\usepackage{subcaption}
\usepackage{float}
\usepackage{tabularx}

\usepackage{eccvabbrv}

\usepackage{graphicx}
\usepackage{booktabs}

\usepackage[accsupp]{axessibility}  

\usepackage{orcidlink}

\newcommand\sotaa{\textcolor{red}}
\newcommand\sotab{\textcolor{blue}}
\definecolor{Gray}{rgb}{0.95, 0.95, 0.95}

\definecolor{cvprblue}{rgb}{0.21,0.49,0.74}
\usepackage[pagebackref,breaklinks,colorlinks,allcolors=cvprblue]{hyperref}


\begin{document}

\title{From Local Windows to Adaptive Candidates via Individualized Exploratory: 
Rethinking Attention for Image Super-Resolution} 

\titlerunning{IET}

\author{
  Chunyu Meng$^{1}$, Wei Long$^{1}$, Shuhang Gu$^{1}$ \\
}

\authorrunning{C. Meng et al.}

\institute{
$^{1}$University of Electronic Science and Technology of China \\
{\tt \small \{mengchunyu88, shuhanggu\}@gmail.com}
}

\maketitle

\begin{figure}[h]
    \centering
    \includegraphics[width=\columnwidth]{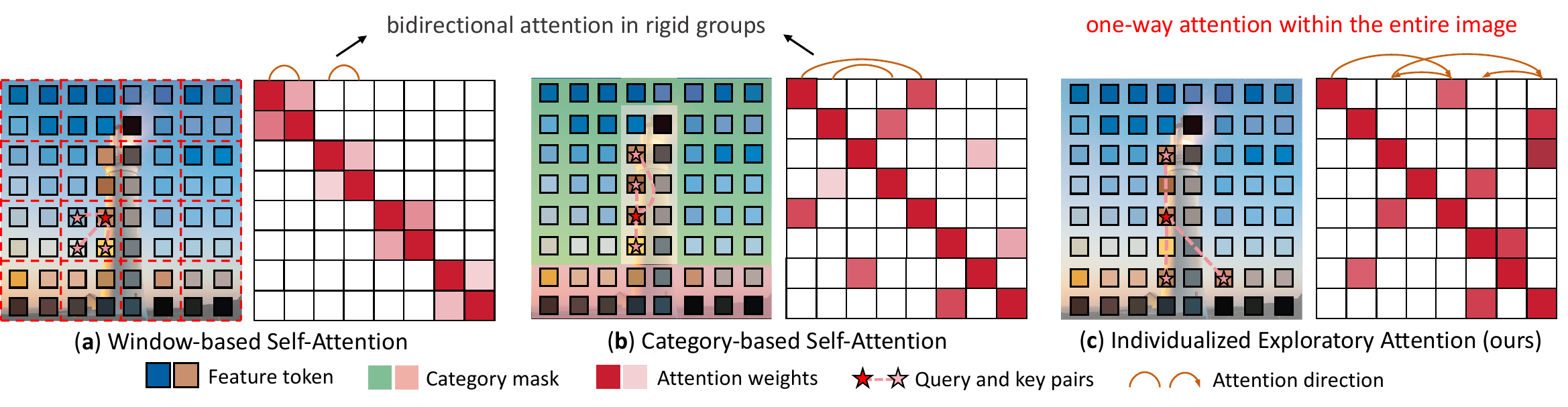}
    \caption{(a) Window-based self-attention groups tokens according to spatial proximity and restricts attention candidates within each window, resulting in attention maps that are symmetric along the diagonal. (b) Category-based self-attention groups tokens based on coarse texture similarity, but still lacks flexibility due to predefined grouping, and similarly produces symmetric attention patterns. (c) Our proposed Individualized Exploratory Attention (IET) allows each token to adaptively and asymmetrically explore one-way neighbors, leading to inherently asymmetric attention maps.}
    \label{fig:comparison_attention}
\end{figure}

\begin{abstract}

Single Image Super-Resolution (SISR) is a fundamental computer vision task that aims to reconstruct a high-resolution (HR) image from a low-resolution (LR) input. Transformer-based methods have achieved remarkable performance by modeling long-range dependencies in degraded images. However, their feature-intensive attention computation incurs high computational cost. To improve efficiency, most existing approaches partition images into fixed groups and restrict attention within each group. Such group-wise attention overlooks the inherent asymmetry in token similarities, thereby failing to enable flexible and token-adaptive attention computation. To address this limitation, we propose the {Individualized Exploratory Transformer (IET)}, which introduces a novel {Individualized Exploratory Attention (IEA)} mechanism that allows each token to adaptively select its own content-aware and independent attention candidates. This token-adaptive and asymmetric design enables more precise information aggregation while maintaining computational efficiency. Extensive experiments on standard SR benchmarks demonstrate that IET achieves state-of-the-art performance under comparable computational complexity.
The code is available at \href{https://github.com/CVL-UESTC/IET.git}{here}.

\end{abstract}

\section{Introduction}
Image Super-Resolution (SR) aims to reconstruct clear and detailed high-resolution (HR) images from low-resolution (LR) inputs. It plays an important role in enhancing perceptual quality and providing reliable visual details for applications such as medical imaging, satellite observation, and video surveillance. However, since one LR image can correspond to multiple possible HR versions, SR remains an ill-posed and challenging task in computer vision.

Earlier SR methods mainly used Convolutional Neural Networks (CNNs)~\cite{dong2014srcnn, Dong_2015_srcnn, gu2015convolutional, Kim_2016_vdsr, lim2017edsr}, which extract features through shared convolutional kernels, and have achieved promising reconstruction results. Recently, Transformer-based methods~\cite{liang2021swinir, chen2023activating, zhang2022elan, zhou2023srformer} have been increasingly applied to image reconstruction, employing a more complex self-attention mechanism that learns feature relationships adaptively instead of fixed kernels. This improvement allows Transformers to capture more flexible and diverse feature representations.

Although the self-attention mechanism in Transformers enables tokens to leverage information from others, its computational complexity grows quadratically with the number of tokens. To efficiently handle the large number of tokens in image data, most methods restrict self-attention computation so that each token interacts with only a limited subset of others. Window-based self-attention is the most common approaches used for this purpose, where an image is divided into local windows, and tokens attend only to the others within the same window~\cite{liu2021swin, chen2022cross, long2025pft}. This design focuses on local feature aggregation but limits the receptive field.
To overcome this limitation, another line of research, known as category-based self-attention, defines attention calculation according to semantic content~\cite{zhang2024atd, liu2025catanet}. Specifically, these methods first categorize all tokens into several classes, then further divide each class into groups, and finally apply attention calculation within each group. Essentially, this strategy reorders the image by grouping semantically similar tokens together, allowing them to mutually assist in reconstruction. These methods consider image content when defining attention, but they still restrict computation within each group, preventing tokens from interacting with others in the same category. Moreover, the accuracy of category boundaries still needs further improvement, as it is difficult to partition tokens into sufficiently fine-grained classes. Significant variation often exists among tokens within the same category, while boundary tokens are unable to interact with more relevant tokens from other categories.

Overall, the above window-based and category-based self-attention methods both fall under grouped attention, which limits the flexibility of individual tokens as they neglect the asymmetric nature of similarity.
In super-resolution, a token may need to pull information from several other tokens to recover fine structures, while those tokens do not need anything in return from the original one.
If this directional relation is not modeled, information flow becomes redundant and the reconstruction is suboptimal.
An ideal SR model should allow each token to flexibly search for its own one-way similar neighbors across a wide spatial range, forming individualized attention candidates that best support feature aggregation.
By contrast, grouped attention imposes rigid group boundaries and encourages symmetric aggregation within groups, which restricts the ability of a token to choose the sources it truly needs.
Modeling asymmetric and individualized relations is thus crucial for restoring complex textures and preserving structures.


In this paper, we propose a novel Individualized Exploratory Attention (IEA) mechanism that connects adjacent attention blocks to progressively select content-aware and asymmetric one-way attention candidates, as shown in Fig.~\ref{fig:comparison_attention}. We leverage the attention map from previous blocks to adjust the similarity relationships: if token A is similar to B, and B is similar to C in the preceding layer, it is very likely that A is similar to C in subsequent layers, and vice versa.
Specifically, in the first block, each token attends only to nearby regions centered on itself. In subsequent blocks, a propagation mechanism allows connection to new long-range similar neighbors, thereby establishing more comprehensive similarity relationships. Meanwhile, a sparsification mechanism is applied to remove low-similarity tokens identified in earlier layers, maintaining an appropriate attention candidates. IEA operates as a dynamic process, starting from a local attention scope and progressively evolving into a content-aware and long-range one through layer-wise refinement. Under comparable computational budgets, our SR model can effectively identify suitable one-way attention candidates over a broader spatial range, leading to improved super-resolution quality.

The main contributions of this paper are as follows:
\begin{itemize}
    \item We propose a Individualized Exploratory  Attention (IEA) mechanism that enables content-aware and token-adaptive attention candidates selection. IEA progressively expands the attention candidates with new similar tokens while pruning low-similarity ones for efficiency.
    \item We develop the Individualized Exploratory Transformer (IET), which utilizes the IEA mechanism to effectively select high-quality attention candidates. As a result, IET can flexibly capture global dependencies without compromising computational efficiency.
    \item Extensive experiments on standard super-resolution benchmarks demonstrate that our method outperforms recent state-of-the-art approaches under comparable computational budgets, and ablation studies validate the effectiveness of each proposed component.
\end{itemize}
\section{Related Work}
\noindent\textbf{Transformer-based SR.}
Over the past decade, deep learning has greatly advanced single-image super-resolution (SR). Starting from SRCNN~\cite{Dong_2015_srcnn}, which introduced deep learning to SR with a simple three-layer CNN, numerous studies have explored various architectural enhancements to boost performance~\cite{Dai_2020_san, gu2019dynamicguidance, Kim_2016_vdsr, kim2016deeply, lim2017edsr, Mei2020image, Mei_2021_nlsa, Niu_2020_han, zhang2018rcan, Zhang_2018_rdn}. VDSR~\cite{Kim_2016_vdsr} deepened the architecture, while DRCN~\cite{kim2016deeply} adopted a recursive design. EDSR~\cite{lim2017edsr} and RDN~\cite{Zhang_2018_rdn} refined residual blocks, further enhancing CNN-based SR. Inspired by Transformers~\cite{vaswani2017attention}, attention mechanisms were introduced to visual tasks—Wang et al.~\cite{wang2018non} first integrated non-local attention into CNNs, demonstrating its effectiveness. Subsequent works, such as CSNLN~\cite{Mei2020image} and NLSA~\cite{Mei_2021_nlsa}, leveraged cross-scale and sparse attention to better capture long-range dependencies while improving computational efficiency.

Subsequently, with the introduction of ViT~\cite{Dosovitskiy_2020_vit} and its variants~\cite{Chu_2021_twin, liu2021swin, Wang_2022_pvt}, the efficacy of pure Transformer-based models in image classification has been established. IPT~\cite{Chen_2020_ipt} first introduces a pre-trained Transformer network to various image restoration tasks, demonstrating the strong representation capability of large-scale Transformers. Then, SwinIR~\cite{liang2021swinir} and CAT~\cite{chen2022cross} employ window-based self-attention mechanisms to model local dependencies efficiently, where SwinIR utilizes shifted windows to aggregate features within a local area, and CAT further improves cross-window interaction through rectangular-window attention and axial shifting. Building on these ideas, HAT~\cite{chen2023activating} enlarges the attention window and incorporates channel attention to extend the receptive field, while PFT~\cite{long2025pft} adopts sparse attention to further expand the interaction range without significantly increasing computational cost. Beyond window-based strategies, ATD~\cite{zhang2024atd} introduces category-based self-attention, which groups tokens with similar semantic characteristics to achieve content-aware feature aggregation. Recently, CATANet~\cite{liu2025catanet} further improves both the feature aggregation process and the handling of category boundaries. In addition, IPG~\cite{Tian2024ipg} improves SR by prioritizing detail-rich pixels to enhance reconstruction.
In this paper, building upon the effectiveness of the attention mechanism in image SR, we propose a IET method that lets tokens to independently explore and select most relevant neighbors from the global space through IEA, thereby enables content-aware and token-adaptive attention candidates selection with low computational cost.

\noindent\textbf{Transformer with Sparse Attention.}
Sparse attention reduces computational complexity by restricting attention candidates. Existing methods typically partition feature maps into rigid groups: for example, SwinIR~\cite{liang2021swinir} groups tokens by spatial windows, ART~\cite{zhang2023accurate} adopts dilated windows, NLSA~\cite{Mei_2021_nlsa} assigns tokens via hash buckets, and ATD~\cite{zhang2024atd} clusters them with learnable centers. In contrast, IEA enables token-adaptive attention candidates. Starting from local neighborhoods, it progressively expands the receptive field through similarity propagation, allowing each token to identify the most relevant neighbors and thereby achieving both accurate and efficient attention computation.

\section{Methodology}

\subsection{Motivation}
\noindent\textbf{Self Attention.} Self-attention~\cite{vaswani2017attention}, the core operation of Transformers, measures token similarity and aggregates features through attention weights. Given $Q, K, V \in \mathbb{R}^{N \times d}$, it can be expressed as
\begin{equation}
A_{sa} = \text{Softmax}\!\left( QK^\top / \sqrt{d} \right), \quad
O_{sa} = A_{sa}V
\label{eq:standard attention}
\end{equation}
where $N$ and $d$ denote the number and dimension of tokens respectively, $A_{sa} \in \mathbb{R}^{N \times N}$ denotes the attention map, and $O_{sa} \in \mathbb{R}^{N \times d}$ denotes the output. Despite its effectiveness, the quadratic computational complexity and redundant interactions among dissimilar tokens significantly limit its efficiency and scalability. For efficiency, grouped attention is commonly adopted, which roughly clusters tokens into groups to improve super-resolution performance. However, it overlooks the asymmetric nature of similarity: token $A$ may need to aggregate information from token $B$, while $B$ does not necessarily require $A$. Therefore, an ideal SR model should adaptively capture such asymmetric relationships and avoid restricting each token within rigid groups.

\noindent\textbf{Individualized Attention.} 
\begin{figure}[t]
\centering

\begin{minipage}[t]{0.48\linewidth}
    \centering
    \includegraphics[width=\linewidth]{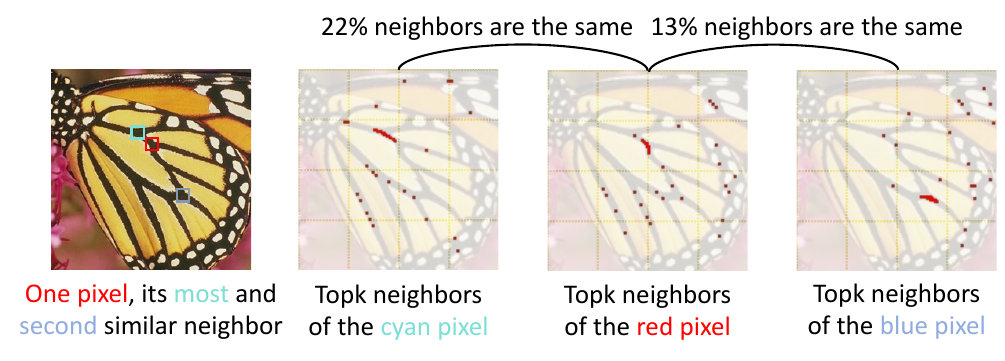}
    \captionof{figure}{Asymmetry and repetition of similarity relationships, the yellow grids represents 32×32 windows.}
    \label{fig:motivation}
\end{minipage}%
\hfill%
\begin{minipage}[t]{0.48\linewidth}
    \centering
    \includegraphics[width=\linewidth]{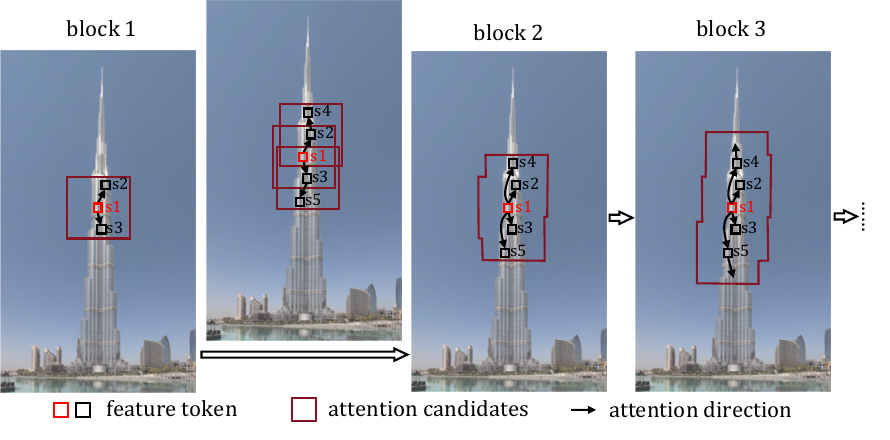}
    \captionof{figure}{The propagation mechanism.}
    \label{fig:propagation}
\end{minipage}

\end{figure}
We analyze token-wise similarity using the attention maps of global self-attention. As illustrated in Fig.~\ref{fig:motivation}, similar neighbors are sparsely distributed over a region far beyond local windows and do not follow fixed spatial priors. This observation is consistent with prior findings: window-based methods often yield suboptimal performance, while enlarging receptive fields via fixed dilated windows brings only limited gains.
Motivated by this insight, we allow tokens to attend to any arbitrary $k$ tokens across the entire feature map, reducing the complexity of global self-attention in Eq.~\ref{eq:standard attention} from $O(N^2)$ to $O(kN)$.
We first introduce the concept of individualized attention, where each token maintains its own asymmetric and independent set of attention candidates. Specifically, individualized attention is implemented via an index matrix $ I \in \mathbb{R}^{N \times k} $, which explicitly records the global indices of the $ k $ most similar neighbors for each token. This formulation yields a flexible attention mechanism: when $ I $ covers all tokens, it reduces to standard self-attention; when restricted to a local window, it becomes window-based self-attention.
Formally,  the individualized attention is proposed as in the following equation
\begin{equation}
A_{ia} = \operatorname{Softmax}( \operatorname{SMM}( Q, K ,I ) / \sqrt{d} ), 
\label{eq:individualized_attention}
\end{equation}
\begin{equation}
O_{ia} = \operatorname{SMM} ( A_{ia}, V ,I ),
\label{eq:individualized_output}
\end{equation}
where SMM denotes sparse matrix multiplication, $A_{ia} \in \mathbb{R}^{N \times k}$ denotes the individualized attention map, and $O_{ia} \in \mathbb{R}^{N \times d}$ denotes the final output. In the individualized attention, the optimization of selecting attention candidates is explicitly simplified into the optimization of the index $I$.

\smallskip
\noindent\textbf{Individualized Exploratory Attention.} To assign token-adaptive and accurate attention candidates, we propose a propagation mechanism. We observe that similar tokens tend to share overlapping neighborhoods, which enables efficient receptive field expansion by exploring higher-order connections in the similarity graph. As illustrated in Fig.~\ref{fig:propagation}, the initial attention candidates are confined to token-centered local neighborhoods. In subsequent blocks, the candidate set progressively propagates toward more correlated regions, resulting in a semantics-aware expansion of the receptive field and allowing each token to identify more suitable bases for reconstruction.

\subsection{Individualized Exploratory Attention}
\begin{figure}[t]
    \centering
    \includegraphics[width=\columnwidth]{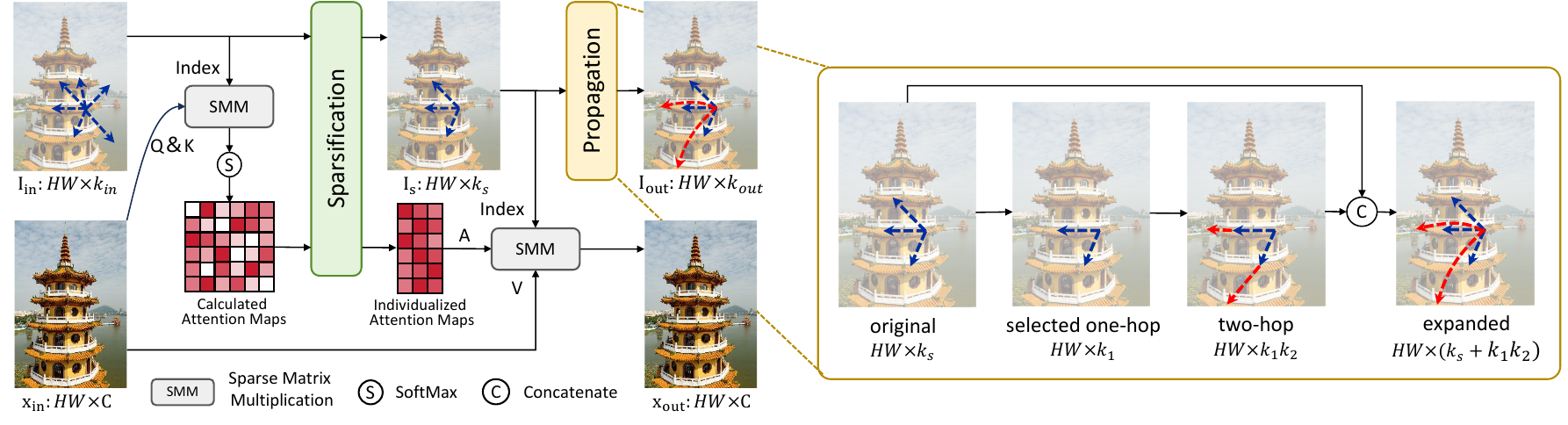}
    \caption{The proposed individualized exploratory attention. We first apply the sparsification mechanism to prune neighbors with low similarity, and then employ the propagation mechanism to explore new two-hop neighbors.}
    \label{fig:individualized_exploratory_attention}
\end{figure}
In IEA, consecutive attention blocks are connected to enable progressive refinement of attention candidates. We propose a propagation mechanism that efficiently enlarges the receptive field by exploring higher-order neighbor relations. To further reduce computational cost, we introduce a sparsification strategy that prunes low-relevance connections and skips unnecessary attention computations. Specifically, the first block initializes with a locally restricted neighborhood, establishing a compact and computationally efficient foundation. In the following blocks, the propagation mechanism promotes second-hop neighbors identified in the previous layer to direct candidates, thereby broadening the attention coverage and capturing longer-range dependencies. Meanwhile, the sparsification mechanism removes connections that have been proved to exhibit low similarity, effectively enhancing the selectivity of attention computation. Through this iterative process of propagation and sparsification across layers, IEA gradually evolves into a content-aware and token-adaptive attention structure, achieving flexible feature aggregation under comparable computational budgets. The following subsections will introduce our attention candidates initialization methods, propagation and sparsification mechanisms in detail and explain how the attention candidates are optimized layer by layer.

\smallskip
\noindent\textbf{Attention Candidates Initialization.} 
As described above, IEA initializes a local attention scope in the first block, and progressively expands it in later blocks. A practical challenge is to ensure that every token discovers two-hop neighbors distinct from its one-hop ones. Following the principles of Expander Graphs, this property holds if each node has a unique neighborhood. To meet this requirement, 
we first consider to define a local attention scope centered on each token. Although this initialization satisfies the requirement, the attention scope of adjacent tokens overlap significantly, leading to a decrease in propagation efficiency. To address this issue, 
we propose a dilation strategy, which performs dense attention computation within a small region while uniformly and sparsely sampling tokens from a larger area. For the distant region, one token are uniformly sampled from each $d \times d$ patch and added to the attention candidates, where $d$ is a predefined dilation factor. This design enables tokens to capture detailed local information as well as coarse global context, thereby achieving a larger initial receptive field.
The choice of $d$ will be explored in ablation studies.

Our initialization strategy shares a superficial similarity with conventional sparse window designs in that it relies on spatial priors. However, unlike traditional window-based methods that treat such spatial grouping as the final attention structure, our design uses it only as a coarse starting point. Although this spatially driven dilation yields limited direct gains, it provides diverse initial candidates that are subsequently refined through semantics-aware propagation, forming a coarse-to-fine construction of token-adaptive attention.

\smallskip
\noindent\textbf{Propagation and Sparsification.} For the subsequent blocks, we introduce Propagation and Sparsification mechanisms to progressively refine the attention candidates. Specifically, given \( Q^{l} \), \( K^{l} \in \mathbb{R}^{N \times D} \) and the initial attention candidates \( I^{l}_{\text{in}} \in \mathbb{R}^{N \times k^{l}_{in}} \), 
we can compute the individualized attention map \( A^{l}_{\text{cal}} \in \mathbb{R}^{N \times k^{l}_{in}} \) according to the individualized attention mechanism as in Eq.\ref{eq:individualized_attention}.
Specifically, we first perform sparsification mechanism on \( A^{l}_{\text{cal}}\) to remove tokens with low attention scores, preserving only those strongly correlated with the query token. For token i, this process is achieved through the following formula:
\begin{equation}
\left\{
\begin{aligned}
S_i &= \operatorname{TopK}(A^{l}_{\text{cal}}[i, :], k^{l}_\text{s}) \\
A^{l}_{\text{s}}[i, ] &= \{\, A^{l}_{\text{cal}}[i, j] \mid j \in S_i \,\}\\
I^{l}_{\text{s}}[i, ] &= \{\, I^{l}_{\text{in}}[i, j] \mid j \in S_i \,\}
\end{aligned}
\right.
\label{eq:select_topk_neighbors_1}
\end{equation}
where $\operatorname{TopK}(\cdot)$ denotes the operation that selects the indices of the top-K elements with the highest values. Then, with \( A^{l}_{\text{s}} \), \( I^{l}_{\text{s}} \in \mathbb{R}^{N \times k^{l}_{s}}\), and \( V^{l} \in \mathbb{R}^{N \times D} \), we can compute the output $O^{l}$ as in Eq.~\ref{eq:individualized_output}. 
After that, our goal is to perform propagation mechanism on \( I^{l}_{\text{s}} \) and \( A^{l}_{\text{s}} \) to obtain more optimal candidates \( I^{l}_{\text{out}} \in \mathbb{R}^{N \times k^{l}_{out}} \), which will then serve as the initial attention candidates \( I^{l+1}_{\text{in}} \) for the next block. Due to the high computational cost of exploring all two-hop neighbors, we adopt a simplified and efficient approximation strategy. For token i, we first select the top \( k^{l}_1 \) one-hop neighbors with the highest attention scores:
\begin{equation}
N_i^{(1)} = \{\, I^{l}_{\text{s}}[i, j] \mid j \in \operatorname{TopK}(A^{l}_{\text{s}}[i, :], k^{l}_1) \,\}
\label{eq:select_topk_neighbors_1}
\end{equation}
where $N_i^{(1)}$ represents the one-hop neighbor set of the token $i$. Then for every selected one-hop neighbor \( u \in N_i^{(1)} \), we still select its top \( k^{l}_2 \) one-hop neighbors:
\begin{equation}
N_u^{(1)} = \{\, I^{l}_{\text{s}}[u, v] \mid v \in \operatorname{TopK}(A^{l}_{\text{s}}[u, :], k^{l}_2) \,\}
\label{eq:select_topk_neighbors_2}
\end{equation}

Next, we can union all the one-hop neighbors of token u to obtain the two-hop neighbors of token i, and finally merge the one- and two-hop neighbors as:
\begin{equation}
N_i^{(2)} = \cup_{u \in N_i^{(1)}} N_u^{(1)}
\label{eq:union}
\end{equation}
\begin{equation}
I^{l}_{\text{out}} = \operatorname{Dedup}\!\left(I^{l}_{\text{s}} \cup N^{(2)}\right)
\label{eq:dedup}
\end{equation}
where $\cup(\cdot)$ denotes the union, and $\operatorname{Dedup}(\cdot)$ denotes the removal of duplicated elements to form a unique set.

The proposed propagation mechanism operates under a locally optimal principle, enlarging the attention scope toward the directions of highest similarity at the current stage. This enables a more comprehensive exploration of potential attention candidates. Meanwhile, the sparsification mechanism prunes low-similarity neighbors, preserving the core relational structure while reducing computational complexity. 
The overall architecture of IEA is illustrated in Fig.~\ref{fig:individualized_exploratory_attention}.
Notably, directly computing the sparse matrix multiplication (SMM) with dense matrix operations is highly inefficient. Thus we modify a CUDA-based sparse matrix multiplication framework~\cite{long2025pft} to suit our setting, enabling efficient individualized attention computation.

\subsection{Similarity-Fused FFN} 
The Feed-Forward Network (FFN) is an important component of Transformer-based architectures, which is traditionally viewed as a token-wise nonlinear mapping. Then several works~\cite{Wang2022uformer, omni_sr, li2021localvit} introduce convolution operations into the FFN to enhance local feature fusion. 
\begin{figure}[t]
    \centering
    \includegraphics[width=\columnwidth]{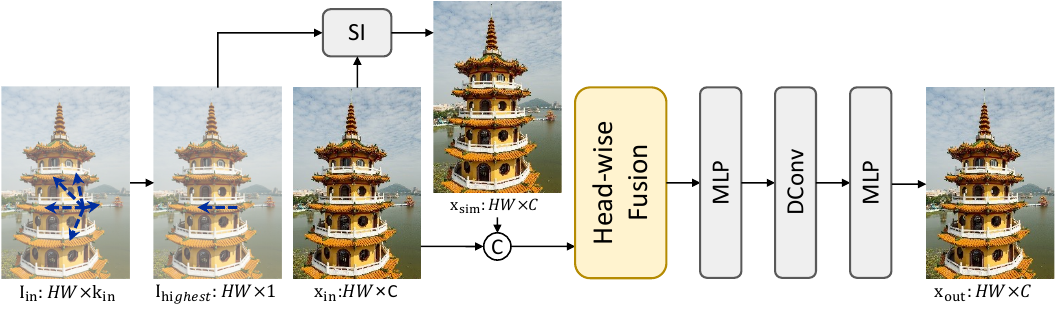}
    \caption{The proposed Similarity-Fused Feed-Forward Network (SF-FFN), which enables cross-channel feature interaction among semantically similar tokens.}
    \label{fig:feed-forward network}
\end{figure}

We further propose the Similarity-Fused Feed-Forward Network (SF-FFN), which exploits the refined token similarities obtained from IEA and enables the fusion of highly correlated tokens. As shown in Fig.~\ref{fig:feed-forward network}, the SF-FFN selects the most similar neighbors $I_{\text{highest}}$ for each token based on the updated neighbor indices $I_{\text{in}}$. It then replaces each token in the input feature map $x_{\text{in}}$ with its most similar neighbor to obtain $x_{\text{sim}}$, and finally fuses the two representations. Specifically, we use a head-wise MLP and depthwise convolution as the fusion module, which introduces only a very small amount of extra computational cost.

From the perspective of the entire Transformer architecture, the SF-FFN provides an important complement to the attention module, whose capacity for cross-channel interaction is inherently limited. In particular, when IEA supplies more comprehensive and accurate similarity relationships, SF-FFN can effectively enhance information exchange among highly correlated tokens.

\subsection{The Overall Network Architecture} The overall architecture of IET follows the design commonly used in recent state-of-the-art Transformer-based SR models~\cite{chen2023activating, zhang2024atd, liang2021swinir}. However, we replace the standard self-attention blocks and feed-forward networks with our proposed IEA block and SF-FFN, respectively Notably, we apply relative positional encoding only in the first block, where the attention scope is still local. For the remaining layers, we utilize LePE positional encoding~\cite{dong2021cswin} directly instead. Moreover, a progressive attention mechanism~\cite{long2025pft} is incorporated into each block to refine and model similarity relationships with greater precision. For classical SR, the IET network consists of 8 blocks, each containing 4 attention layers. It adopts multi-head attention with 6 heads and uses a total of 240 channels. For lightweight SR, the IET-light network also consists of 8 blocks, each containing only 3 layers. It uses 3 attention heads with 54 total channels. For all models, we only perform propagation mechanism in the last layer of first four blocks, with \( k_1\) being [22, 20, 14, 12], \( k_2\) being [12, 10, 7, 6] respectively.

\section{Experiments}

\subsection{Experiment Setting} We train our model on the DF2K dataset, which is constructed by merging DIV2K~\cite{timofte2017div2k} and Flickr2K~\cite{lim2017edsr}. The model is optimized for 500K iterations using the Muon~\cite{jordan2024muon} optimizer with an initial learning rate of $2 \times 10^{-4}$. The model is initially trained on image patches of size $50 \times 50$ with a batch size of 60, and subsequently fine-tuned using $75 \times 75$ patches with a reduced batch size of 30. Moreover, we employ a MultistepLR scheduler, which reduces the learning rate by half at specified iterations [250000, 400000, 450000, 475000]. Furthermore, all computational cost measurements reported in this paper are based on outputs with a spatial resolution of $1280 \times 640$.

\subsection{Ablation Study} We conduct ablation studies on the proposed IET-light model, with all models trained for 250k iterations on the DIV2K dataset at ×4 scale.

\smallskip
\noindent\textbf{Effects of Propagation and Sparsification Mechanism.}
In order to demonstrate the effectiveness of the key design choices in the proposed individualized exploratory transformer (IET) model, we establish five models with different component combinations and evaluate their performance on Urban100 and Manga109 datasets, as shown in Table~\ref{tab:ablation_components}.
The first row represents the baseline model, where all modules are disabled. This configuration is equivalent to dilated attention, meaning that all modules use the same fixed attention candidate.

\begin{table}[t]
\centering
\label{tab:ablation_all}

\begin{minipage}[t]{0.48\linewidth}
\centering
\setlength{\tabcolsep}{2.5pt}
\captionof{table}{Ablation study on the effects of each component.}
\label{tab:ablation_components}
\resizebox{\linewidth}{!}{
\begin{tabular}{c|c|c|c|c}
\hline
\textbf{Propagation} & \textbf{Sparsification} & \textbf{SF-FFN} & \textbf{FLOPs} & \textbf{PSNR (dB)} \\
\hline
 & & & 180.4G & 26.17 / 30.58 \\
 & \checkmark & & 61.2G & 26.28 / 30.66 \\
 \checkmark & & & 215.9G & 26.91 / 30.92 \\
 \checkmark & \checkmark & & 68.9G & 26.96 / 31.05 \\
 \checkmark & \checkmark & \checkmark & 69.5G & \textbf{27.03 / 31.09} \\
\hline
\end{tabular}
}
\end{minipage}
\hfill
\begin{minipage}[t]{0.48\linewidth}
\centering
\setlength{\tabcolsep}{2pt}
\renewcommand{\arraystretch}{0.95}
\captionof{table}{Ablation on the number of prppagation steps.}
\label{tab:expansion_steps}
\resizebox{\linewidth}{!}{
\begin{tabular}{c|c|c|c|c}
\hline
\textbf{Propagation Depth} & \textbf{FLOPs} & \textbf{Set5} & \textbf{Urban100} & \textbf{Manga109} \\
\hline
1 & 65.4G & 32.29 & 26.52 & 30.70 \\
2 & 68.5G & 32.35 & 26.66 & 30.81 \\
3 & 69.1G & 32.48 & 26.83 & 30.97 \\
4 & 69.5G & \textbf{32.55} & \textbf{27.03} & \textbf{31.09} \\
5 & 69.8G & 32.53 & 26.96 & 31.05 \\
\hline
\end{tabular}
}
\end{minipage}

\end{table}

\begin{table}[t]
\centering
\label{tab:ablation_all}

\begin{minipage}[t]{0.48\linewidth}
\centering
\setlength{\tabcolsep}{2.5pt}
\captionof{table}{Ablation on dilation settings.}
\label{tab:dilation}
\resizebox{\linewidth}{!}{
\begin{tabular}{c|ccccc}
\hline
$d$ for Train & 1 & 2 & 3 & 2 & 2 \\
$d$ for Infer & 1 & 2 & 3 & 3 & 4 \\
\hline
PSNR (dB) & 26.86 & 26.97 & 26.99 & 27.03 & 27.02 \\
\hline
\end{tabular}
}
\end{minipage}
\hfill
\begin{minipage}[t]{0.48\linewidth}
\centering
\setlength{\tabcolsep}{2pt}
\renewcommand{\arraystretch}{0.95}
\captionof{table}{Ablation on the ratio of $k_1$ to $k_2$ in propagation, evaluated on Urban100.}
\label{tab:k1k2_ratio}
\resizebox{0.85\linewidth}{!}{
\begin{tabular}{c|cccc}
\hline
\textbf{$k_1$/$k_2$} & 0.5 & 1 & 2 & 3 \\ 
\hline
PSNR(dB) & 26.98 & 27.01 & 27.03 & 27.00 \\ 
\hline
\end{tabular}
}
\end{minipage}

\end{table}

\begin{table}[t]
\centering
\label{tab:ablation_all}

\begin{minipage}[t]{0.48\linewidth}
\centering
\setlength{\tabcolsep}{2.5pt}
\captionof{table}{Ablation on SF-FFN placement across the final blocks. 0 block from last denotes that there is no SF-FFN.}
\label{tab:sf_ffn_placement}
\resizebox{\linewidth}{!}{
\begin{tabular}{c|cccc}
\hline
Blocks (from last) & 0 & 1 & 2 & 3 \\
\hline
PSNR  & 26.96 & 27.00 & 27.03 & 27.02 \\
\hline
\end{tabular}
}
\end{minipage}
\hfill
\begin{minipage}[t]{0.48\linewidth}
\centering
\setlength{\tabcolsep}{2pt}
\renewcommand{\arraystretch}{0.95}
\captionof{table}{Ablation on the number of similar tokens fused in SF-FFN. 0 similar token denotes that there is no SF-FFN.}
\label{tab:sf_ffn_neighbors}
\resizebox{\linewidth}{!}{
\begin{tabular}{c|ccc}
\hline
Fused Similar Tokens & 0 & 1 & 2 \\
\hline
PSNR & 26.96 & 27.03 & 26.97 \\
\hline
\end{tabular}
}
\end{minipage}

\end{table}
The second model, which enables the sparsification mechanism, shows a marginal improvement of about 0.11 dB on Urban100 and 0.08 dB on Manga109. Skipping low-correlation attention computations not only reduces computational cost but also improves performance. This suggests that spatially initialized attention candidates contain a large number of weakly related neighbors, which are not only computationally redundant but can even degrade performance.
The third model introduces the propagation mechanism, achieving a significant improvement of about 0.66 dB on Urban100 and 0.29 dB on Manga109. The propagation mechanism not only compensates for the coarse nature of the initial dilation-based expansion but also further extends the receptive field. Simply introducing direct connections to two-hop neighbors brings considerable performance improvement, highlighting the importance of relational propagation beyond local neighborhoods.
The fourth model, combining the propagation and sparsification, improves PSNR by 0.79 on Urban10 and 0.47 dB on Manga109 compared with the baseline.
The final model integrates SF-FFN based on the fourth one, achieving an improvement of 0.07 dB on Urban100 and 0.04 dB on Manga109. SF-FFN concatenates similar tokens along the channel dimension before projection, thereby introducing cross-channel interactions. This design effectively complements the attention mechanism, which lacks explicit cross-channel modeling capability.

Importantly, the primary performance gain comes from the propagation mechanism. While dilated initialization expands attention candidates based solely on spatial proximity, it remains a coarse expansion strategy. Its role is mainly to provide sufficiently diverse initial neighbors for subsequent relational exploration. In contrast, propagation performs semantic-level expansion by exploring neighbors of neighbors, effectively introducing meaningful long-range dependencies. This relational reasoning over two-hop connections accounts for the major improvement.
Sparsification brings a modest performance gain, but its principal benefit lies in reducing computational redundancy by removing weakly correlated connections. Moreover, the SF-FFN introduces explicit cross-channel interactions to further improve the model.

\noindent\textbf{Effects of Propagation Depth.} We apply the propagation mechanism only in the last layer of each block to ensure stability. To study how many blocks should incorporate the propagation mechanism, we evaluate models with different numbers of propagation steps applied in the earlier blocks. As shown in Tab.~\ref{tab:expansion_steps}, using four propagation steps achieves the best performance. This result indicates that propagation is effective for exploring new neighbors and enriching attention candidates. Meanwhile, applying propagation too late or too frequently may introduce unstable or noisy long-range candidates, suggesting that later blocks require higher accuracy in similarity modeling and benefit less from excessive propagation.

\smallskip
\noindent\textbf{Effects of Dilation on Attention Candidates Initialization.}
In IET, the attention candidate initialization uniformly samples tokens from each $d \times d$ patch, where the dilation factor $d$ determines the spatial sampling density. Different dilations can be used during training and inference since dilation only affects attention candidate selection, not model parameters. In general, a larger $d$ corresponds to a wider initial receptive field , allowing tokens to access more distant contextual information, while $d=1$ restricts attention to the local neighborhood. As shown in Tab.~\ref{tab:dilation}, increasing $d$ during both training and inference effectively broadens the receptive field and enhances feature aggregation, leading to improved reconstruction quality. However, beyond $d=3$, the performance gain saturates, suggesting that excessively long-range dependencies contribute marginally to SR recovery.
Moreover, during training, using $d=2$ achieves slightly better results than $d=3$, as a smaller dilation allows for smaller patches and thus larger batch sizes, leading to more stable optimization. This experiment also verifies that the similarity modeling in attention is weakly correlated with spatial distance, highlighting the rationality of selecting attention candidates according to image content rather than fixed spatial positions. Finally, we adopt $d=2$ for training and $d=3$ for inference in our IET model.

\smallskip
\noindent\textbf{Effects of $k_1$/$k_2$ Ratio on Propagation.} 
In our IEA, tokens first selects their $k_1$ most similar token as one-hop neighbors, and then collects $k_2$ most similar tokens for each one-hop neighbor to obtain two-hop neighbors. We further study the impact of the ratio between $k_1$ and $k_2$ on reconstruction performance. As shown in Tab.~\ref{tab:k1k2_ratio}, the model achieves the best results when $k_1/k_2=2$, suggesting that appropriately enlarging the one-hop neighbor set improves stability and enhances the effectiveness of the propagation.

\begin{table}[t]
\captionsetup{font={small}}
\scriptsize
\caption{Quantitative comparison (PSNR/SSIM) with state-of-the-art methods on \textbf{classical SR} task. The best and second best results are colored with \sotaa{red} and \sotab{blue}. Results on ×3 model are presented in the supplementary material.}
\vspace{-6mm}
\label{tab: quantitative comparison classical}
  \begin{center}
      
  \resizebox{\textwidth}{!}{
  \begin{tabular}{|p{2.3cm}|c|c|c|cc|cc|cc|cc|cc|}
    \hline
    \multirow{2}{*}{\textbf{Method}} & \multirow{2}{*}{\textbf{Scale}} & \multirow{2}{*}{\textbf{Params}} & \multirow{2}{*}{\textbf{FLOPs}} & \multicolumn{2}{c|}{\textbf{Set5}} & \multicolumn{2}{c|}{\textbf{Set14}} & \multicolumn{2}{c|}{\textbf{BSD100}} & \multicolumn{2}{c|}{\textbf{Urban100}} & \multicolumn{2}{c|}{\textbf{Manga109}} \\

    & & & & PSNR & SSIM & PSNR & SSIM & PSNR & SSIM & PSNR & SSIM & PSNR & SSIM   \\

    \hline

    EDSR~\cite{lim2017edsr}         & $\times$2 & 42.6M & 22.14T & 38.11 & 0.9602 & 33.92 & 0.9195 & 32.32 & 0.9013 & 32.93 & 0.9351 & 39.10 & 0.9773 \\
    RCAN~\cite{zhang2018rcan}       & $\times$2 & 15.4M & 7.02T & 38.27 & 0.9614 & 34.12 & 0.9216 & 32.41 & 0.9027 & 33.34 & 0.9384 & 39.44 & 0.9786 \\
    HAN~\cite{Niu_2020_han}         & $\times$2 & 63.6M & 7.24T & 38.27 & 0.9614 & 34.16 & 0.9217 & 32.41 & 0.9027 & 33.35 & 0.9385 & 39.46 & 0.9785 \\
    IPT~\cite{Chen_2020_ipt}        & $\times$2 & 115M & 7.38T  & 38.37 & - & 34.43 & - & 32.48 & - & 33.76 & - & - & - \\
    SwinIR~\cite{liang2021swinir}   & $\times$2 & 11.8M & 3.04T & 38.42 & 0.9623 & 34.46 & 0.9250 & 32.53 & 0.9041 & 33.81 & 0.9433 & 39.92 & 0.9797 \\
    CAT-A~\cite{chen2022cross}      & $\times$2 & 16.5M & 5.08T & 38.51 & 0.9626 & 34.78 & 0.9265 & 32.59 & 0.9047 & 34.26 & 0.9440 & 40.10 & 0.9805 \\
    ART~\cite{zhang2023accurate}    & $\times$2 & 16.4M & 7.04T & 38.56 & 0.9629 & 34.59 & 0.9267 & 32.58 & 0.9048 & 34.30 & 0.9452 & 40.24 & 0.9808 \\
    HAT~\cite{chen2023activating}   & $\times$2 & 20.6M & 5.81T & 38.63 & 0.9630 & 34.86 & 0.9274 & 32.62 & 0.9053 & 34.45 & 0.9466 & 40.26 & 0.9809 \\
    MambaIRv2-B~\cite{guo2024mambairv2}  & $\times$2 & 22.9M & 6.27T & 38.65 & 0.9631 & 34.89 & 0.9275 & 32.62 & 0.9053 & 34.49 & 0.9468 & 40.42 & 0.9810 \\
    IPG~\cite{Tian2024ipg}   & $\times$2 & 18.1M & 5.35T & 38.61 & 0.9632 & 34.47 & 0.9270 & 32.60 & 0.9052 & 34.48 & 0.9464 & 40.24 & 0.9810 \\
    ATD~\cite{zhang2024atd}   & $\times$2 & 20.1M & 6.07T & 38.61 & 0.9629 & 34.95 & 0.9276 & 32.65 & 0.9056 & 34.70 & 0.9476 & 40.37 & 0.9810 \\
    PFT~\cite{long2025pft}   & $\times$2 & 19.6M & 5.03T & \sotab{38.68} & \sotab{0.9635} & \sotab{35.00} & \sotab{0.9280} & \sotab{32.67} & \sotab{0.9058} & \sotab{34.90} & \sotab{0.9490} & \sotab{40.49} & \sotab{0.9815} \\

    \rowcolor{Gray}
    \textbf{IET} (ours)              & $\times$2 & 19.7M & 5.02T & \sotaa{38.74} & \sotaa{0.9636} & \sotaa{35.10} & \sotaa{0.9286} & \sotaa{32.71} & \sotaa{0.9064} & \sotaa{35.09} & \sotaa{0.9500} & \sotaa{40.60} & \sotaa{0.9816} \\

    \midrule
    \hline

    EDSR~\cite{lim2017edsr}         & $\times$3 & 43.0M & 9.82T & 34.65 & 0.9280 & 30.52 & 0.8462 & 29.25 & 0.8093 & 28.80 & 0.8653 & 34.17 & 0.9476 \\
    RCAN~\cite{zhang2018rcan}       & $\times$3 & 15.6M & 3.12T & 34.74 & 0.9299 & 30.65 & 0.8482 & 29.32 & 0.8111 & 29.09 & 0.8702 & 34.44 & 0.9499 \\
    HAN~\cite{Niu_2020_han}         & $\times$3 & 64.2M & 3.21T & 34.75 & 0.9299 & 30.67 & 0.8483 & 29.32 & 0.8110 & 29.10 & 0.8705 & 34.48 & 0.9500 \\
    IPT~\cite{Chen_2020_ipt}        & $\times$3 & 116M & 3.28T & 34.81 & -      & 30.85 & -      & 29.38 & -      & 29.49 & -      & -     & -      \\
    SwinIR~\cite{liang2021swinir}   & $\times$3 & 11.9M & 1.35T & 34.97 & 0.9318 & 30.93 & 0.8534 & 29.46 & 0.8145 & 29.75 & 0.8826 & 35.12 & 0.9537 \\
    CAT-A~\cite{chen2022cross}      & $\times$3 & 16.6M & 2.26T & 35.06 & 0.9326 & 31.04 & 0.8538 & 29.52 & 0.8160 & 30.12 & 0.8862 & 35.38 & 0.9546 \\
    ART~\cite{zhang2023accurate}    & $\times$3 & 16.6M & 3.12T & 35.07 & 0.9325 & 31.02 & 0.8541 & 29.51 & 0.8159 & 30.10 & 0.8871 & 35.39 & 0.9548 \\
    HAT~\cite{chen2023activating}   & $\times$3 & 20.8M & 2.58T & 35.07 & 0.9329 & 31.08 & 0.8555 & 29.54 & 0.8167 & 30.23 & 0.8896 & 35.53 & 0.9552 \\
    MambaIRv2-B~\cite{guo2024mambairv2}    & $\times$3 & 23.1M & 2.78T & 35.18 & 0.9334 & 31.12 & 0.8557 & 29.55 & 0.8169 & 30.28 & 0.8905 & 35.61 & 0.9556 \\
    IPG~\cite{Tian2024ipg}   & $\times$3 & 18.3M & 2.39T & 35.10 & 0.9332 & 31.10 & 0.8554 & 29.53 & 0.8168 & 30.36 & 0.8901 & 35.53 & 0.9554 \\
    ATD~\cite{zhang2024atd}    & $\times$3 & 20.3M & 2.69T & 35.11 & 0.9330 & 31.13 & 0.8556 & 29.57 & 0.8176 & 30.46 & 0.8917 & 35.63 & 0.9558 \\
    PFT~\cite{long2025pft}   & $\times$3 & 19.8M & 2.23T & \sotab{35.15} & \sotab{0.9333} & \sotab{31.16} & \sotab{0.8561} & \sotab{29.58} & \sotab{0.8178} & \sotab{30.56} & \sotab{0.8931} & \sotab{35.67} & \sotab{0.9560} \\

    \rowcolor{Gray}
    \textbf{IET} (ours)             & $\times$3 & 19.9M & 2.25T & \sotaa{35.20} & \sotaa{0.9337} & \sotaa{31.23} & \sotaa{0.8571} & \sotaa{29.61} & \sotaa{0.8185} & \sotaa{30.81} & \sotaa{0.8966} & \sotaa{35.82} & \sotaa{0.9566} \\

    \hline \hline
    
    EDSR~\cite{lim2017edsr}         & $\times$4 & 43.0M & 5.54T & 32.46 & 0.8968 & 28.80 & 0.7876 & 27.71 & 0.7420 & 26.64 & 0.8033 & 31.02 & 0.9148 \\
    RCAN~\cite{zhang2018rcan}       & $\times$4 & 15.6M & 1.76T & 32.63 & 0.9002 & 28.87 & 0.7889 & 27.77 & 0.7436 & 26.82 & 0.8087 & 31.22 & 0.9173 \\
    HAN~\cite{Niu_2020_han}         & $\times$4 & 64.2M & 1.81T & 32.64 & 0.9002 & 28.90 & 0.7890 & 27.80 & 0.7442 & 26.85 & 0.8094 & 31.42 & 0.9177 \\
    IPT~\cite{Chen_2020_ipt}        & $\times$4 & 116M & 1.85T  & 32.64 & -      & 29.01 & -      & 27.82 & -      & 27.26 & -      & -     & -      \\   
    SwinIR~\cite{liang2021swinir}   & $\times$4 & 11.9M & 0.76T & 32.92 & 0.9044 & 29.09 & 0.7950 & 27.92 & 0.7489 & 27.45 & 0.8254 & 32.03 & 0.9260 \\
    CAT-A~\cite{chen2022cross}      & $\times$4 & 16.6M & 1.27T & 33.08 & 0.9052 & 29.18 & 0.7960 & 27.99 & 0.7510 & 27.89 & 0.8339 & 32.39 & 0.9285 \\
    ART~\cite{zhang2023accurate}    & $\times$4 & 16.6M & 1.76T & 33.04 & 0.9051 & 29.16 & 0.7958 & 27.97 & 0.7510 & 27.77 & 0.8321 & 32.31 & 0.9283 \\
    HAT~\cite{chen2023activating}   & $\times$4 & 20.8M & 1.45T & 33.04 & 0.9056 & 29.23 & 0.7973 & 28.00 & 0.7517 & 27.97 & 0.8368 & 32.48 & 0.9292 \\
    MambaIRv2-B~\cite{guo2024mambairv2}   & $\times$4 & 23.1M & 1.57T & 33.14 & 0.9057 & 29.23 & 0.7975 & 28.00 & 0.7511 & 27.89 & 0.8344 & 32.57 & 0.9295 \\
    IPG~\cite{Tian2024ipg}   & $\times$4 & 17.0M & 1.30T & 33.15 & 0.9062 & 29.24 & 0.7973 & 27.99 & 0.7519 & 28.13 & 0.8392 & 32.53 & 0.9300 \\
    ATD~\cite{zhang2024atd}   & $\times$4 & 20.3M & 1.52T & 33.10 & 0.9058 & 29.24 & 0.7974 & 28.01 & 0.7526 & 28.17 & 0.8404 & 32.62 & 0.9306 \\
    PFT~\cite{long2025pft}   & $\times$4 & 19.8M & 1.26T & \sotab{33.15} & \sotab{0.9065} & \sotab{29.29} & \sotab{0.7978} & \sotab{28.02} & \sotab{0.7527} & \sotab{28.20} & \sotab{0.8412} & \sotab{32.63} & \sotab{0.9306} \\
    
    \rowcolor{Gray}
    \textbf{IET} (ours)             & $\times$4 & 19.8M & 1.26T & \sotaa{33.22} & \sotaa{0.9069} & \sotaa{29.35} & \sotaa{0.7994} & \sotaa{28.06} & \sotaa{0.7536} & \sotaa{28.43} & \sotaa{0.8464} & \sotaa{32.81} & \sotaa{0.9318} \\
    \hline
  \end{tabular}
  }
  \end{center}

\end{table}

\smallskip
\noindent\textbf{Effects of SF-FFN.} Our network consists of 8 Transformer blocks, and we explore the effect of inserting 
the proposed SF-FFN at different positions. 
As shown in Tab.~\ref{tab:sf_ffn_placement}, placing SF-FFN in the early blocks yields limited improvement since the attention maps at early stages are relatively noisy and cannot accurately represent token similarity. In contrast, inserting SF-FFN in the later blocks allows it to operate on more stable attention patterns, 
leading to better reconstruction performance. The best results are achieved 
when SF-FFN is placed after the last two blocks. we conclude that attention maps in the later blocks represent more accurate similarity relationships, making them more suitable for cross-channel fusion in SF-FFN.
We further analyze how many similar tokens should be fused in the SF-FFN. 
As shown in Tab.~\ref{tab:sf_ffn_neighbors}, when each token interacts with 
only its most similar neighbor, the model achieves the best trade-off between 
performance and efficiency. Using more neighbors introduces redundant information 
and slightly degrades reconstruction accuracy.
From these results,  we draw a conclusion that in the SR task, where high-precision similarity is critical, each token benefits most from cross-channel interaction with only its single most similar neighbor.

\begin{table}[t]
\captionsetup{font={small}}
\scriptsize
  
\caption{Quantitative comparison (PSNR/SSIM) with state-of-the-art methods on \textbf{lightweight SR} task. The best and second best results are colored with \sotaa{red} and \sotab{blue}. Results on ×3 model are presented in the supplementary material.}
\vspace{-6mm}
\label{tab: quantitative comparison light}
  \begin{center}
      
  \resizebox{\textwidth}{!}{
  \begin{tabular}{|p{2.7cm}|c|c|c|cc|cc|cc|cc|cc|}
    \hline
    \multirow{2}{*}{\textbf{Method}} & \multirow{2}{*}{\textbf{Scale}} & \multirow{2}{*}{\textbf{Params}} & \multirow{2}{*}{\textbf{FLOPs}} & \multicolumn{2}{c|}{\textbf{Set5}} & \multicolumn{2}{c|}{\textbf{Set14}} & \multicolumn{2}{c|}{\textbf{BSD100}} & \multicolumn{2}{c|}{\textbf{Urban100}} & \multicolumn{2}{c|}{\textbf{Manga109}} \\

    & & & & PSNR & SSIM & PSNR & SSIM & PSNR & SSIM & PSNR & SSIM & PSNR & SSIM   \\

    \hline \hline
    CARN~\cite{Ahn_2018_carn}               & $\times$2 & 1,592K & 222.8G & 37.76 & 0.9590 & 33.52 & 0.9166 & 32.09 & 0.8978 & 31.92 & 0.9256 & 38.36 & 0.9765 \\
    IMDN~\cite{Hui_2019_imdn}               & $\times$2 & 694K & 158.8G   & 38.00 & 0.9605 & 33.63 & 0.9177 & 32.19 & 0.8996 & 32.17 & 0.9283 & 38.88 & 0.9774 \\
    LAPAR-A~\cite{Li_2020_lapar}            & $\times$2 & 548K & 171G   & 38.01 & 0.9605 & 33.62 & 0.9183 & 32.19 & 0.8999 & 32.10 & 0.9283 & 38.67 & 0.9772 \\
    LatticeNet~\cite{Luo_2020_latticenet}   & $\times$2 & 756K & 169.5G   & 38.15 & 0.9610 & 33.78 & 0.9193 & 32.25 & 0.9005 & 32.43 & 0.9302 & -     & -      \\
    SwinIR-light~\cite{liang2021swinir}     & $\times$2 & 910K & 244G   & 38.14 & 0.9611 & 33.86 & 0.9206 & 32.31 & 0.9012 & 32.76 & 0.9340 & 39.12 & 0.9783 \\
    ELAN~\cite{zhang2022elan}          & $\times$2 & 582K & 203G   & 38.17 & 0.9611 & 33.94 & 0.9207 & 32.30 & 0.9012 & 32.76 & 0.9340 & 39.11 & 0.9782 \\
    SwinIR-NG~\cite{Choi_2022_swinirng}     & $\times$2 & 1181K & 274.1G  & 38.17 & 0.9612 & 33.94 & 0.9205 & 32.31 & 0.9013 & 32.78 & 0.9340 & 39.20 & 0.9781 \\
    OmniSR~\cite{omni_sr}                   & $\times$2 & 772K & 194.5G   & 38.22 & 0.9613 & 33.98 & 0.9210 & 32.36 & 0.9020 & 33.05 & 0.9363 & 39.28 & 0.9784 \\
    MambaIRv2-light~\cite{guo2024mambairv2}          & $\times$2 & 774K & 286.3G   & 38.26 & 0.9615 & 34.09 & 0.9221 & 32.36 & 0.9019 & 33.26 & 0.9378 & 39.35 & 0.9785 \\
    IPG-Tiny~\cite{Tian2024ipg}                & $\times$2 & 872K & 245.2G  & 38.27 & 0.9616 & \sotab{34.24} & \sotab{0.9236} & 32.35 & 0.9018 & 33.04 & 0.9359 & 39.31 & 0.9786 \\
    ATD-light~\cite{zhang2024atd}                & $\times$2 & 753K & 348.6G  & 38.28 & 0.9616 & 34.11 & 0.9217 & 32.39 & 0.9023 & 33.27 & 0.9376 & 39.51 & 0.9789 \\
    PFT-light~\cite{long2025pft}                   & $\times$2 & 776K & 278.3G   & \sotab{38.36} & \sotab{0.9620} & 34.19 & 0.9232 & \sotab{32.43} & \sotab{0.9030} & \sotab{33.67} & \sotab{0.9411} & \sotab{39.55} & \sotab{0.9792} \\
    
    
    \rowcolor{Gray}
    \textbf{IET-light} (Ours)      & $\times$2 & 783K & 277.4G  & \sotaa{38.44} & \sotaa{0.9624} & \sotaa{34.28} & \sotaa{0.9245} & \sotaa{32.50} & \sotaa{0.9038} & \sotaa{34.03} & \sotaa{0.9435} & \sotaa{39.75} & \sotaa{0.9794} \\

    \midrule
    \hline

    CARN~\cite{Ahn_2018_carn}               & $\times$3 & 1,592K & 118.8G & 34.29 & 0.9255 & 30.29 & 0.8407 & 29.06 & 0.8034 & 28.06 & 0.8493 & 33.50 & 0.9440 \\
    IMDN~\cite{Hui_2019_imdn}               & $\times$3 & 703K & 71.5G   & 34.36 & 0.9270 & 30.32 & 0.8417 & 29.09 & 0.8046 & 28.17 & 0.8519 & 33.61 & 0.9445 \\
    LAPAR-A~\cite{Li_2020_lapar}            & $\times$3 & 544K & 114G   & 34.36 & 0.9267 & 30.34 & 0.8421 & 29.11 & 0.8054 & 28.15 & 0.8523 & 33.51 & 0.9441 \\
    LatticeNet~\cite{Luo_2020_latticenet}   & $\times$3 & 765K & 76.3G   & 34.53 & 0.9281 & 30.39 & 0.8424 & 29.15 & 0.8059 & 28.33 & 0.8538 & -     & -      \\
    SwinIR-light~\cite{liang2021swinir}     & $\times$3 & 918K & 111G   & 34.62 & 0.9289 & 30.54 & 0.8463 & 29.20 & 0.8082 & 28.66 & 0.8624 & 33.98 & 0.9478 \\
    ELAN~\cite{zhang2022elan}          & $\times$3 & 590K & 90.1G   & 34.61 & 0.9288 & 30.55 & 0.8463 & 29.21 & 0.8081 & 28.69 & 0.8624 & 34.00 & 0.9478 \\
    SwinIR-NG~\cite{Choi_2022_swinirng}     & $\times$3 & 1190K & 114.1G  & 34.64 & 0.9293 & {30.58} & {0.8471} & 29.24 & 0.8090 & 28.75 & 0.8639 & {34.22} & {0.9488} \\
    OmniSR~\cite{omni_sr}                   & $\times$3 & 780K & 88.4G   & 34.70 & 0.9294 & 30.57 & 0.8469 & 29.28 & 0.8094 & 28.84 & 0.8656 & {34.22} & 0.9487 \\
    MambaIRv2-light~\cite{guo2024mambairv2}          & $\times$2 & 781K & 126.7G   & 34.71 & 0.9298 & 30.68 & 0.8483 & 29.26 & 0.8098 & 29.01 & 0.8689 & 34.41 & 0.9497 \\
    IPG-Tiny~\cite{Tian2024ipg}          & $\times$2 & 878K & 109.0G   & 34.64 & 0.9292 & 30.61 & 0.8470 & 29.26 & 0.8097 & 28.93 & 0.8666 & 34.30 & 0.9493 \\
    ATD-light~\cite{zhang2024atd}                & $\times$2 & 753K & 154.7G  & 34.70 & 0.9300 & 30.68 & 0.8485 & 29.32 & 0.8109 & 29.16 & 0.8710 & \sotab{34.60} & 0.9505 \\
    PFT-light~\cite{long2025pft}                   & $\times$2 & 776K & 123.5G   & \sotab{34.81} & \sotab{0.9305} & \sotab{30.75} & \sotab{0.8493} & \sotab{29.33} & \sotab{0.8116} & \sotab{29.43} & \sotab{0.8759} & \sotab{34.60} & \sotab{0.9510} \\
    
    \rowcolor{Gray}
    \textbf{IET-light} (Ours)               & $\times$3 & 790K & 123.1G  & \sotaa{34.90} & \sotaa{0.9314} & \sotaa{30.83} & \sotaa{0.8504} & \sotaa{29.41} & \sotaa{0.8135} & \sotaa{29.75} & \sotaa{0.8808} & \sotaa{34.89} & \sotaa{0.9524} \\

    \hline \hline
    CARN~\cite{Ahn_2018_carn}               & $\times$4 & 1,592K & 90.9G & 32.13 & 0.8937 & 28.60 & 0.7806 & 27.58 & 0.7349 & 26.07 & 0.7837 & 30.47 & 0.9084 \\
    IMDN~\cite{Hui_2019_imdn}               & $\times$4 & 715K & 40.9G  & 32.21 & 0.8948 & 28.58 & 0.7811 & 27.56 & 0.7353 & 26.04 & 0.7838 & 30.45 & 0.9075 \\
    LAPAR-A~\cite{Li_2020_lapar}            & $\times$4 & 659K & 94G   & 32.15 & 0.8944 & 28.61 & 0.7818 & 27.61 & 0.7366 & 26.14 & 0.7871 & 30.42 & 0.9074 \\
    LatticeNet~\cite{Luo_2020_latticenet}   & $\times$4 & 777K & 43.6G   & 32.30 & 0.8962 & 28.68 & 0.7830 & 27.62 & 0.7367 & 26.25 & 0.7873 & -     & -      \\
    SwinIR-light~\cite{liang2021swinir}     & $\times$4 & 930K & 63.6G   & 32.44 & 0.8976 & 28.77 & 0.7858 & 27.69 & 0.7406 & 26.47 & 0.7980 & 30.92 & 0.9151 \\
    ELAN~\cite{zhang2022elan}          & $\times$4 & 582K & 54.1G   & 32.43 & 0.8975 & 28.78 & 0.7858 & 27.69 & 0.7406 & 26.54 & 0.7982 & 30.92 & 0.9150 \\
    SwinIR-NG~\cite{Choi_2022_swinirng}     & $\times$4 & 1201K & 63G  & 32.44 & 0.8980 & 28.83 & 0.7870 & 27.73 & 0.7418 & 26.61 & 0.8010 & 31.09 & 0.9161 \\
    OmniSR~\cite{omni_sr}                   & $\times$4 & 792K & 50.9G   & 32.49 & 0.8988 & 28.78 & 0.7859 & 27.71 & 0.7415 & 26.65 & 0.8018 & 31.02 & 0.9151 \\
    IPG-Tiny~\cite{Tian2024ipg}                   & $\times$4 & 887K & 61.3G   & 32.51 & 0.8987 & 28.85 & 0.7873 & 27.73 & 0.7418 & 26.78 & 0.8050 & 31.22 & 0.9176 \\
    MambaIRv2-light~\cite{guo2024mambairv2}          & $\times$2 & 790K & 75.6G   & 32.51 & 0.8992 & 28.84 & 0.7878 & 27.75 & 0.7426 & 26.82 & 0.8079 & 31.24 & 0.9182 \\
    ATD-light~\cite{zhang2024atd}                & $\times$2 & 769K & 87.1G  & 32.62 & 0.8997 & 28.87 & 0.7884 & 27.77 & 0.7439 & 26.97 & 0.8107 & 31.47 & 0.9198 \\
    PFT-light~\cite{long2025pft}                   & $\times$2 & 792K & 69.6G   & \sotab{32.63} & \sotab{0.9005} & \sotab{28.92} & \sotab{0.7891} & \sotab{27.79} & \sotab{0.7445} & \sotab{27.20} & \sotab{0.8171} & \sotab{31.51} & \sotab{0.9204} \\
    

    \rowcolor{Gray}
    \textbf{IET-light} (Ours)      & $\times$4 & 801K & 69.4G   & \sotaa{32.70} & \sotaa{0.9019} & \sotaa{29.00} & \sotaa{0.7908} & \sotaa{27.85} & \sotaa{0.7464} & \sotaa{27.41} & \sotaa{0.8227} & \sotaa{31.73} & \sotaa{0.9226} \\
    \hline
  \end{tabular}
  }
  \end{center}
\end{table}

\subsection{Comparison with State-of-the-Art Methods}
We evaluate our model against various super-resolution baselines on standard benchmark datasets, including Set5~\cite{Bevilacqua2012set5}, Set14~\cite{Zeyde_2012_set14}, BSD100~\cite{Martin_2002_BSD100}, Urban100~\cite{Huang_2015_Urban100}, and Manga109~\cite{Matsui_2016_Manga109}. The comparison covers both traditional and recent advanced SR approaches, such as EDSR~\cite{lim2017edsr}, RCAN~\cite{zhang2018rcan}, HAN~\cite{Niu_2020_han}, SwinIR~\cite{liang2021swinir}, CAT~\cite{chen2022cross}, ART~\cite{zhang2023accurate}, HAT~\cite{chen2023activating}, IPG~\cite{Tian2024ipg}, ATD~\cite{zhang2024atd}, and PFT~\cite{long2025pft}. The performance of all models is measured using PSNR and SSIM under $\times2$, $\times3$, and $\times4$ upscaling settings. Computational costs are evaluated at an output resolution of $1280 \times 640$.

The results shown in Table~\ref{tab: quantitative comparison classical}, indicate that with a similar number of parameters, the proposed IET model achieves significantly better performance than PFT. Notably, on the ×2 Urban100 benchmark, IET outperforms PFT by \textbf{0.19dB} and ATD by \textbf{0.39dB}.
For lightweight super-resolution, we further compare our method with efficient SR networks such as CARN~\cite{Ahn_2018_carn}, IMDN~\cite{Hui_2019_imdn}, LAPAR~\cite{Li_2020_lapar}, SwinIR~\cite{liang2021swinir}, ELAN~\cite{zhang2022elan}, and OmniSR~\cite{omni_sr}. As shown in Table~\ref{tab: quantitative comparison light}, the proposed IET-light consistently surpasses PFT-light~\cite{long2025pft} on all benchmark datasets. Specifically, on the $\times2$ Urban100 dataset, IET-light exceeds PFT-light by \textbf{0.36dB} and ATD-light by \textbf{0.76dB}. Moreover, IET-light outperforms SwinIR by 0.02dB on Set5 and 0.22dB on Urban100, while requiring only \textbf{9.1\%} of the total computational complexity. For the $\times4$ SR setting, IET-light also achieves a \textbf{0.21dB} improvement over PFT-light and a \textbf{0.44dB} gain over ATD-light on Urban100. The strong performance of IET stems from the proposed Individualized Exploratory Attention, which enables content-aware and token-adaptive attention candidates, and then aggregates information among the tokens that are most appropriate for feature enhancement.

\begin{figure}[t]
    \centering
    \includegraphics[width=\textwidth]{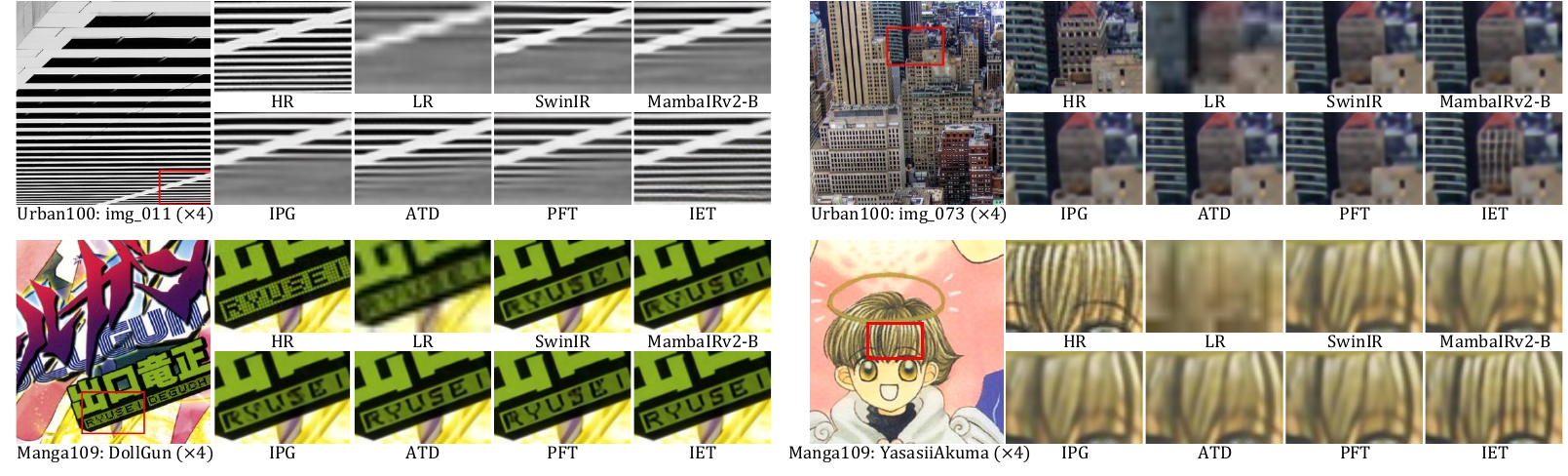}
    \caption{Visual comparisons of \textbf{IET} and other SOTA image super-resolution methods.}
    \label{fig:visual_comparison}
\end{figure}

We also provide some visual examples using different methods to qualitatively verify the efficacy of IET, as shown in Fig.\ref{fig:visual_comparison}, which clearly demonstrate our advantage in recovering sharp edges and clean textures.

\subsection{Visualization Analysis} 
\begin{figure}[t]
    \centering
    \includegraphics[width=\textwidth]{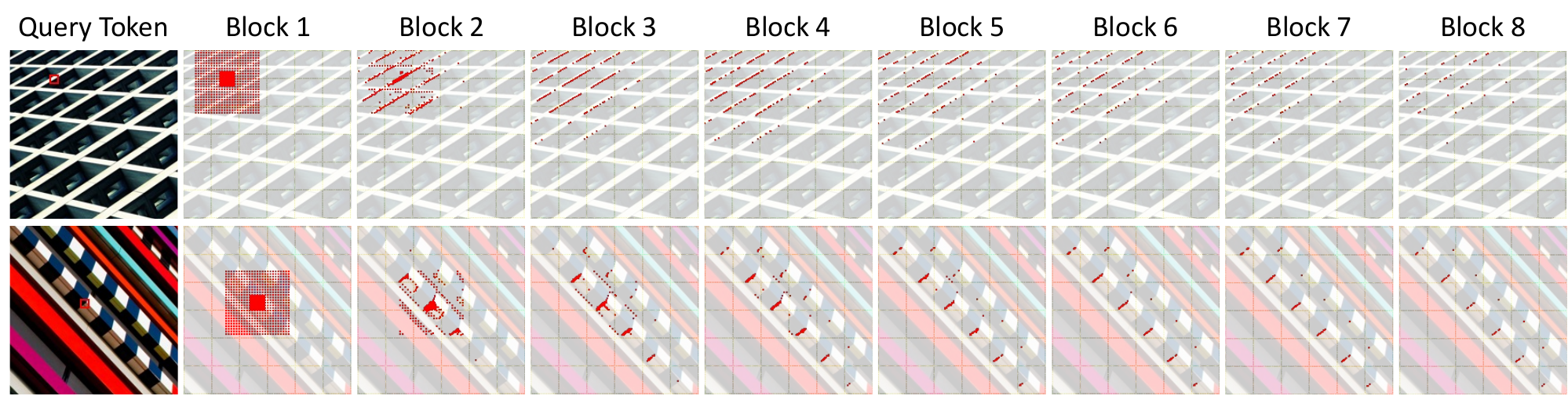}
    \caption{Visualization of the attention candidates initialization and propagation process. The yellow grids represents 32×32 windows, which is the largest window size used among window-based methods.}
    \label{fig:attention_candidates_visualization}
\end{figure}
We further visualize the refinement process of attention candidates in Fig.~\ref{fig:attention_candidates_visualization}, where yellow grids represent 32×32 windows and red points denotes attention candidates. In the first block, the attention scope is initialized as a dense local and sparse global region centered on each token. This allows every token to accurately capture nearby information while roughly perceiving distant context, providing rich initial cues for subsequent propagation. In the following blocks, IEA gradually expands the attention candidates by aggregate with new two-hop similar tokens while pruning low-similarity neighbors, maintaining efficiency with comparable computational cost.

\section{Conclusion} We propose the Individualized Exploratory Transformer (IET), a content-aware self-attention framework with token-adaptive candidate selection. Its core module refines attention through propagation and sparsification, modeling asymmetric similarity while reducing redundancy. This flexible and efficient design achieves state-of-the-art results on super-resolution benchmarks and shows potential for broader vision and language tasks.

\section*{Acknowledgement}  This work was supported by the National Natural Science
Foundation of China (No. 62476051) and the Sichuan
Natural Science Foundation (No. 2024NSFTD0041).

%
%
\bibliographystyle{splncs04}
\bibliography{main}

@String(CVPR= {IEEE Conf. Comput. Vis. Pattern Recog.})

@String(ICCV= {Int. Conf. Comput. Vis.})

@String(ECCV= {Eur. Conf. Comput. Vis.})

@String(ICLR = {Int. Conf. Learn. Represent.})

@String(CVPRW= {IEEE Conf. Comput. Vis. Pattern Recog. Worksh.})

@String(CVPR  = {CVPR})

@String(ICCV  = {ICCV})

@String(ECCV  = {ECCV})

@String(ICLR  = {ICLR})

@String(CVPRW= {CVPRW})

@InProceedings{chen2023activating,
    author    = {Chen, Xiangyu and Wang, Xintao and Zhou, Jiantao and Qiao, Yu and Dong, Chao},
    title     = {Activating More Pixels in Image Super-Resolution Transformer},
    booktitle = {Proceedings of the IEEE/CVF Conference on Computer Vision and Pattern Recognition (CVPR)},
    month     = {June},
    year      = {2023},
    pages     = {22367-22377}
}

@misc{liu2021swin,
      title={Swin Transformer: Hierarchical Vision Transformer using Shifted Windows}, 
      author={Ze Liu and Yutong Lin and Yue Cao and Han Hu and Yixuan Wei and Zheng Zhang and Stephen Lin and Baining Guo},
      year={2021},
      eprint={2103.14030},
      archivePrefix={arXiv},
      primaryClass={cs.CV}
}

@misc{liang2021swinir,
      title={SwinIR: Image Restoration Using Swin Transformer}, 
      author={Jingyun Liang and Jiezhang Cao and Guolei Sun and Kai Zhang and Luc Van Gool and Radu Timofte},
      year={2021},
      eprint={2108.10257},
      archivePrefix={arXiv},
      primaryClass={eess.IV}
}

@misc{dong2021cswin,
      title={CSWin Transformer: A General Vision Transformer Backbone with Cross-Shaped Windows}, 
        author={Xiaoyi Dong and Jianmin Bao and Dongdong Chen and Weiming Zhang and Nenghai Yu and Lu Yuan and Dong Chen and Baining Guo},
        year={2021},
        eprint={2107.00652},
        archivePrefix={arXiv},
        primaryClass={cs.CV}
}

@misc{vaswani2017attention,
      title={Attention Is All You Need}, 
      author={Ashish Vaswani and Noam Shazeer and Niki Parmar and Jakob Uszkoreit and Llion Jones and Aidan N. Gomez and Lukasz Kaiser and Illia Polosukhin},
      year={2017},
      eprint={1706.03762},
      archivePrefix={arXiv},
      primaryClass={cs.CL}
}

@inbook{zhang2018rcan,   title={Image Super-Resolution Using Very Deep Residual Channel Attention Networks},  url={http://dx.doi.org/10.1007/978-3-030-01234-2_18},  DOI={10.1007/978-3-030-01234-2_18},  booktitle={Computer Vision – ECCV 2018,Lecture Notes in Computer Science},  author={Zhang, Yulun and Li, Kunpeng and Li, Kai and Wang, Lichen and Zhong, Bineng and Fu, Yun},  year={2018},  month={Oct},  pages={294–310},  language={en-US}  }

@inproceedings{timofte2017div2k,   title={NTIRE 2017 Challenge on Single Image Super-Resolution: Methods and Results},  url={http://dx.doi.org/10.1109/cvprw.2017.149},  DOI={10.1109/cvprw.2017.149},  booktitle={2017 IEEE Conference on Computer Vision and Pattern Recognition Workshops (CVPRW)},  author={Timofte, Radu and Agustsson, Eirikur and Gool, Luc Van and Yang, Ming-Hsuan and Zhang, Lei and Lim, Bee and Son, Sanghyun and Kim, Heewon and Nah, Seungjun and Lee, Kyoung Mu and others},  year={2017},  month={Aug},  language={en-US}  }

@inproceedings{lim2017edsr,   title={Enhanced Deep Residual Networks for Single Image Super-Resolution},  url={http://dx.doi.org/10.1109/cvprw.2017.151},  DOI={10.1109/cvprw.2017.151},  booktitle={2017 IEEE Conference on Computer Vision and Pattern Recognition Workshops (CVPRW)},  author={Lim, Bee and Son, Sanghyun and Kim, Heewon and Nah, Seungjun and Lee, Kyoung Mu},  year={2017},  month={Aug},  language={en-US}  }

@inproceedings{Bevilacqua2012set5,   title={Low-complexity single-image super-resolution based on nonnegative neighbor embedding},  url={http://dx.doi.org/10.5244/c.26.135},  DOI={10.5244/c.26.135},  booktitle={Procedings of the British Machine Vision Conference 2012},  author={Bevilacqua, Marco and Roumy, Aline and Guillemot, Christine and Morel, Marie-line Alberi},  year={2012},  month={Sep},  language={en-US}  }

@inbook{Zeyde_2012_set14,   title={On Single Image Scale-Up Using Sparse-Representations},  url={http://dx.doi.org/10.1007/978-3-642-27413-8_47},  DOI={10.1007/978-3-642-27413-8_47},  booktitle={Curves and Surfaces,Lecture Notes in Computer Science},  author={Zeyde, Roman and Elad, Michael and Protter, Matan},  year={2012},  month={Jan},  pages={711–730},  language={en-US}  }

@inproceedings{Martin_2002_BSD100,   title={A database of human segmented natural images and its application to evaluating segmentation algorithms and measuring ecological statistics},  url={http://dx.doi.org/10.1109/iccv.2001.937655},  DOI={10.1109/iccv.2001.937655},  booktitle={Proceedings Eighth IEEE International Conference on Computer Vision. ICCV 2001},  author={Martin, D. and Fowlkes, C. and Tal, D. and Malik, J.},  year={2002},  month={Nov},  language={en-US}  }

@misc{jordan2024muon,
  author       = {Keller Jordan and Yuchen Jin and Vlado Boza and You Jiacheng and
                  Franz Cesista and Laker Newhouse and Jeremy Bernstein},
  title        = {Muon: An optimizer for hidden layers in neural networks},
  year         = {2024},
  url          = {https://kellerjordan.github.io/posts/muon/}
}

@inproceedings{Huang_2015_Urban100,   title={Single image super-resolution from transformed self-exemplars},  url={http://dx.doi.org/10.1109/cvpr.2015.7299156},  DOI={10.1109/cvpr.2015.7299156},  booktitle={2015 IEEE Conference on Computer Vision and Pattern Recognition (CVPR)},  author={Huang, Jia-Bin and Singh, Abhishek and Ahuja, Narendra},  year={2015},  month={Oct},  language={en-US}  }

@article{Matsui_2016_Manga109,   title={Sketch-based manga retrieval using manga109 dataset},  url={http://dx.doi.org/10.1007/s11042-016-4020-z},  DOI={10.1007/s11042-016-4020-z},  journal={Multimedia Tools and Applications},  author={Matsui, Yusuke and Ito, Kota and Aramaki, Yuji and Fujimoto, Azuma and Ogawa, Toru and Yamasaki, Toshihiko and Aizawa, Kiyoharu},  year={2016},  month={Nov},  pages={21811–21838},  language={en}  }

@misc{Ahn_2018_carn,   title={Fast, Accurate, and Lightweight Super-Resolution with Cascading Residual Network},  journal={arXiv: Computer Vision and Pattern Recognition},  author={Ahn, Namhyuk and Kang, Byungkon and Sohn, Kyung-Ah},  year={2018},  month={Mar},  language={en-US}  }

@article{Dong_2015_srcnn,   title={Image Super-Resolution Using Deep Convolutional Networks},  url={http://dx.doi.org/10.1109/tpami.2015.2439281},  DOI={10.1109/tpami.2015.2439281},  journal={IEEE Transactions on Pattern Analysis and Machine Intelligence},  author={Dong, Chao and Loy, Chen Change and He, Kaiming and Tang, Xiaoou},  year={2015},  month={Jun},  pages={295–307},  language={en-US}  }

@inproceedings{Kim_2016_vdsr,   title={Accurate Image Super-Resolution Using Very Deep Convolutional Networks},  url={http://dx.doi.org/10.1109/cvpr.2016.182},  DOI={10.1109/cvpr.2016.182},  booktitle={2016 IEEE Conference on Computer Vision and Pattern Recognition (CVPR)},  author={Kim, Jiwon and Lee, Jung Kwon and Lee, Kyoung Mu},  year={2016},  month={Dec},  language={en-US}  }

@inproceedings{Dai_2020_san,   title={Second-Order Attention Network for Single Image Super-Resolution},  url={http://dx.doi.org/10.1109/cvpr.2019.01132},  DOI={10.1109/cvpr.2019.01132},  booktitle={2019 IEEE/CVF Conference on Computer Vision and Pattern Recognition (CVPR)},  author={Dai, Tao and Cai, Jianrui and Zhang, Yongbing and Xia, Shu-Tao and Zhang, Lei},  year={2020},  month={Jan},  language={en-US}  }

@inbook{Niu_2020_han,   title={Single Image Super-Resolution via a Holistic Attention Network},  url={http://dx.doi.org/10.1007/978-3-030-58610-2_12},  DOI={10.1007/978-3-030-58610-2_12},  booktitle={Computer Vision – ECCV 2020,Lecture Notes in Computer Science},  author={Niu, Ben and Wen, Weilei and Ren, Wenqi and Zhang, Xiangde and Yang, Lianping and Wang, Shuzhen and Zhang, Kaihao and Cao, Xiaochun and Shen, Haifeng},  year={2020},  month={Oct},  pages={191–207},  language={en-US}  }

@inproceedings{Mei_2021_nlsa,   title={Image Super-Resolution with Non-Local Sparse Attention},  url={http://dx.doi.org/10.1109/cvpr46437.2021.00352},  DOI={10.1109/cvpr46437.2021.00352},  booktitle={2021 IEEE/CVF Conference on Computer Vision and Pattern Recognition (CVPR)},  author={Mei, Yiqun and Fan, Yuchen and Zhou, Yuqian},  year={2021},  month={Nov},  language={en-US}  }

@misc{Zhang_2018_rdn,   title={Residual Dense Network for Image Super-Resolution},  journal={Cornell University - arXiv},  author={Zhang, Yulun and Tian, Yapeng and Kong, Yu and Zhong, Bineng and Fu, Yun},  year={2018},  month={Feb},  language={en-US}  }

@misc{Dosovitskiy_2020_vit,   title={An Image is Worth 16x16 Words: Transformers for Image Recognition at Scale},  journal={arXiv: Computer Vision and Pattern Recognition},  author={Dosovitskiy, Alexey and Beyer, Lucas and Kolesnikov, Alexander and Weissenborn, Dirk and Zhai, Xiaohua and Unterthiner, Thomas and Dehghani, Mostafa and Minderer, Matthias and Heigold, Georg and Gelly, Sylvain and Uszkoreit, Jakob and Houlsby, Neil},  year={2020},  month={Oct},  language={en-US}  }

@misc{Chu_2021_twin,   title={Twins: Revisiting the Design of Spatial Attention in Vision Transformers},  journal={Neural Information Processing Systems},  author={Chu, Xiangxiang and Tian, Zhi and Wang, Yuqing and Zhang, Bo and Ren, Haibing and Wei, Xiaolin and Xia, Huaxia and Shen, Chunhua},  year={2021},  month={Dec},  language={en-US}  }

@inproceedings{Wang_2022_pvt,   title={Pyramid Vision Transformer: A Versatile Backbone for Dense Prediction without Convolutions},  url={http://dx.doi.org/10.1109/iccv48922.2021.00061},  DOI={10.1109/iccv48922.2021.00061},  booktitle={2021 IEEE/CVF International Conference on Computer Vision (ICCV)},  author={Wang, Wenhai and Xie, Enze and Li, Xiang and Fan, Deng-Ping and Song, Kaitao and Liang, Ding and Lu, Tong and Luo, Ping and Shao, Ling},  year={2022},  month={Mar},  language={en-US}  }

@inproceedings{zhang2022elan,
  title={Efficient Long-Range Attention Network for Image Super-Resolution},
  author={Zhang, Xindong and Zeng, Hui and Guo, Shi and Zhang, Lei},
  booktitle={European Conference on Computer Vision},
  pages={649--667},
  publisher={Springer},
  year={2022}
}

@inproceedings{zhou2023srformer,
  title={SRFormer: Permuted Self-Attention for Single Image Super-Resolution},
  author={Zhou, Yupeng and Li, Zhen and Guo, Chun-Le and Bai, Song and Cheng, Ming-Ming and Hou, Qibin},
  booktitle={Proceedings of the IEEE/CVF Conference on Computer Vision and Pattern Recognition},
  pages={12780--12791},
  year={2023}
}

@inproceedings{Hui_2019_imdn,   title={Lightweight Image Super-Resolution with Information Multi-distillation Network},  url={http://dx.doi.org/10.1145/3343031.3351084},  DOI={10.1145/3343031.3351084},  booktitle={Proceedings of the 27th ACM International Conference on Multimedia},  author={Hui, Zheng and Gao, Xinbo and Yang, Yunchu and Wang, Xiumei},  year={2019},  month={Oct},  language={en-US}  }

@misc{Li_2020_lapar,   title={LAPAR: Linearly-Assembled Pixel-Adaptive Regression Network for Single Image Super-resolution and Beyond},  journal={Neural Information Processing Systems},  author={Li, Wenbo and Zhou, Kun and Qi, Lu and Jiang, Nianjuan and Lu, Jiangbo and Jia, Jiaya},  year={2020},  month={Jan},  language={en-US}  }

@inbook{Luo_2020_latticenet,   title={LatticeNet: Towards Lightweight Image Super-Resolution with Lattice Block},  url={http://dx.doi.org/10.1007/978-3-030-58542-6_17},  DOI={10.1007/978-3-030-58542-6_17},  booktitle={Computer Vision – ECCV 2020,Lecture Notes in Computer Science},  author={Luo, Xiaotong and Xie, Yuan and Zhang, Yulun and Qu, Yanyun and Li, Cuihua and Fu, Yun},  year={2020},  month={Nov},  pages={272–289},  language={en-US}  }

@misc{Choi_2022_swinirng,   title={N-Gram in Swin Transformers for Efficient Lightweight Image Super-Resolution},  author={Choi, Haram and Lee, Jeongmin and Yang, Jihoon},  year={2022},  month={Nov},  language={en-US}  }

@misc{Chen_2020_ipt,   title={Pre-Trained Image Processing Transformer},  journal={Cornell University - arXiv},  author={Chen, Hanting and Wang, Yunhe and Guo, Tianyu and Xu, Chang and Deng, Yiping and Liu, Zhenhua and Ma, Siwei and Xu, Chunjing and Xu, Chao and Gao, Wen},  year={2020},  month={Dec},  language={en-US}  }

@inproceedings{Mei2020image,
  title={Image Super-Resolution with Cross-Scale Non-Local Attention and Exhaustive Self-Exemplars Mining},
  author={Mei, Yiqun and Fan, Yuchen and Zhou, Yuqian and Huang, Lichao and Huang, Thomas S and Shi, Humphrey},
  booktitle={Proceedings of the IEEE Conference on Computer Vision and Pattern Recognition (CVPR)},
  year={2020}
}

@inproceedings{kim2016deeply,
  title={Deeply-recursive convolutional network for image super-resolution},
  author={Kim, Jiwon and Lee, Jung Kwon and Lee, Kyoung Mu},
  booktitle={Proceedings of the IEEE conference on computer vision and pattern recognition},
  pages={1637--1645},
  year={2016}
}

@article{long2025pft,
  title={Progressive Focused Transformer for Single Image Super-Resolution},
  author={Long, Wei and Zhou, Xingyu and Zhang, Leheng and Gu, Shuhang},
  journal={arXiv preprint arXiv:2503.20337},
  year={2025}
}

@InProceedings{zhang2024atd,
    author    = {Zhang, Leheng and Li, Yawei and Zhou, Xingyu and Zhao, Xiaorui and Gu, Shuhang},
    title     = {Transcending the Limit of Local Window: Advanced Super-Resolution Transformer with Adaptive Token Dictionary},
    booktitle = {Proceedings of the IEEE/CVF Conference on Computer Vision and Pattern Recognition (CVPR)},
    month     = {June},
    year      = {2024},
    pages     = {2856-2865}
}

@article{gu2019dynamicguidance,
  title={Learned Dynamic Guidance for Depth Image Reconstruction},
  author={Gu, Shuhang and Guo, Shi and Zuo, Wangmeng and Chen, Yunjin and Timofte, Radu and Van Gool, Luc and Zhang, Lei},
  journal={IEEE Transactions on Pattern Analysis and Machine Intelligence},
  volume={42},
  number={10},
  pages={2437--2452},
  year={2019}
}

@inproceedings{liu2025catanet,
  title={CATANet: Efficient Content-Aware Token Aggregation for Lightweight Image Super-Resolution},
  author={Liu, Xin and Liu, Jie and Tang, Jie and Wu, Gangshan},
  booktitle={Proceedings of the IEEE/CVF Conference on Computer Vision and Pattern Recognition},
  pages={17902--17912},
  year={2025}
}

@article{guo2024mambairv2,
  title={MambaIRv2: Attentive State Space Restoration},
  author={Guo, Hang and Guo, Yong and Zha, Yaohua and Zhang, Yulun and Li, Wenbo and Dai, Tao and Xia, Shu-Tao and Li, Yawei},
  journal={arXiv preprint arXiv:2411.15269},
  year={2024}
}

@InProceedings{Tian2024ipg,
    author    = {Tian, Yuchuan and Chen, Hanting and Xu, Chao and Wang, Yunhe},
    title     = {Image Processing GNN: Breaking Rigidity in Super-Resolution},
    booktitle = {Proceedings of the IEEE/CVF Conference on Computer Vision and Pattern Recognition (CVPR)},
    month     = {June},
    year      = {2024},
    pages     = {24108-24117}
}

@inproceedings{chen2022cross,
    title={Cross Aggregation Transformer for Image Restoration},
    author={Chen, Zheng and Zhang, Yulun and Gu, Jinjin and Zhang, Yongbing and Kong, Linghe and Yuan, Xin},
    booktitle={NeurIPS},
    year={2022}
}

@inproceedings{omni_sr,
  title      = {Omni Aggregation Networks for Lightweight Image Super-Resolution},
  author     = {Wang, Hang and Chen, Xuanhong and Ni, Bingbing and Liu, Yutian and Liu jinfan},
  booktitle  = {Conference on Computer Vision and Pattern Recognition},
  year       = {2023}
}

@inproceedings{loshchilov2018decoupled,
  title={Decoupled Weight Decay Regularization},
  author={Loshchilov, Ilya and Hutter, Frank},
  booktitle={International Conference on Learning Representations},
  year={2018}
}

@article{li2021localvit,
  title={LocalViT: Bringing Locality to Vision Transformers},
  author={Li, Yawei and Zhang, Kai and Cao, Jiezhang and Timofte, Radu and Van Gool, Luc},
  journal={arXiv preprint arXiv:2104.05707},
  year={2021}
}

@inproceedings{zhang2023accurate,
  title={Accurate Image Restoration with Attention Retractable Transformer},
  author={Zhang, Jiale and Zhang, Yulun and Gu, Jinjin and Zhang, Yongbing and Kong, Linghe and Yuan, Xin},
  booktitle={ICLR},
  year={2023}
}

@InProceedings{Wang2022uformer,
    author    = {Wang, Zhendong and Cun, Xiaodong and Bao, Jianmin and Zhou, Wengang and Liu, Jianzhuang and Li, Houqiang},
    title     = {Uformer: A General U-Shaped Transformer for Image Restoration},
    booktitle = {Proceedings of the IEEE/CVF Conference on Computer Vision and Pattern Recognition (CVPR)},
    month     = {June},
    year      = {2022},
    pages     = {17683-17693}
}

@inproceedings{wang2018non,
  title={Non-local neural networks},
  author={Wang, Xiaolong and Girshick, Ross and Gupta, Abhinav and He, Kaiming},
  booktitle={Proceedings of the IEEE conference on computer vision and pattern recognition},
  pages={7794--7803},
  year={2018}
}

@inproceedings{gu2015convolutional,
  title={Convolutional sparse coding for image super-resolution},
  author={Gu, Shuhang and Zuo, Wangmeng and Xie, Qi and Meng, Deyu and Feng, Xiangchu and Zhang, Lei},
  booktitle={Proceedings of the IEEE International Conference on Computer Vision},
  pages={1823--1831},
  year={2015}
}

@inproceedings{dong2014srcnn,
  title={Learning a deep convolutional network for image super-resolution},
  author={Dong, Chao and Loy, Chen Change and He, Kaiming and Tang, Xiaoou},
  booktitle={Computer Vision--ECCV 2014: 13th European Conference, Zurich, Switzerland, September 6-12, 2014, Proceedings, Part IV},
  pages={184--199},
  year={2014},
  organization={Springer}
}

\clearpage
\setcounter{page}{1}
\renewcommand{\thesection}{\Alph{section}}

\begin{center}
{\large\bfseries From Local Windows to Adaptive Candidates via Individualized Exploratory: 
Rethinking Attention for Image Super-Resolution}

{\bfseries Supplementary Material}
\end{center}

In this supplementary material, we provide additional details on model training, inference time efficiency comparisons, and more comprehensive visual results. Specifically, in Section A, we present the training details for the IET and IET-light models. Subsequently, in Section B, we compare the inference time efficiency of different models. Finally, in Section C, we provide more detailed visualizations of the model’s results.

\section{Training Details}
\textbf{IET.} We follow prior works~\cite{chen2023activating, liang2021swinir} and adopt the DF2K dataset, which merges DIV2K~\cite{timofte2017div2k} and Flickr2K~\cite{lim2017edsr}, as our training corpus. The training of IET is performed in two stages.
In the first stage, we randomly crop $50 \times 50$ LR patches and their corresponding HR counterparts, using a batch size of~60 and setting the IEA dilation to~2. We jointly employ the Muon~\cite{jordan2024muon} and AdamW~\cite{loshchilov2018decoupled} optimizers ($\beta_1=0.9$, $\beta_2=0.99$), minimizing the $\ell_1$ pixel loss. Convolution kernels are optimized with Muon (learning rate $1 \times 10^{-3}$), while linear projection layers use AdamW (learning rate $2 \times 10^{-4}$). Both learning rates are halved at the 250k iteration milestone, and this stage runs for 300k iterations for the $\times 2$ model.
In the second stage, we increase the IEA dilation to~3 and enlarge the patch size to $75 \times 75$. Training proceeds for another 250k iterations with a batch size of~30. Muon is applied to convolution kernels (learning rate $1.5 \times 10^{-4}$) and AdamW to linear projection layers (learning rate $3 \times 10^{-5}$), with learning rates halved at $[100\text{k}, 150\text{k}, 170\text{k}, 190\text{k}, 200\text{k}]$. For the $\times 3$ and $\times 4$ settings, we skip the first stage and finetune directly from the pretrained $\times 2$ model.
A warm-up schedule is adopted at the beginning of each stage, gradually increasing the learning rate from zero to its initialized value to ensure a smooth and stable optimization process.

\smallskip
\noindent\textbf{IET-light.} To ensure fair comparisons with previous state-of-the-art methods, we use only the DIV2K dataset for training. Following IET and other prior works, we train the $\times 2$ model from scratch and finetune the $\times 3$ and $\times 4$ models from the pretrained $\times 2$ backbone. Specifically, the $\times 2$ IET-light model is trained for 400k iterations, while the $\times 3$ and $\times 4$ variants are finetuned for 100k iterations. During training, we randomly crop $75 \times 75$ low-resolution patches with their HR counterparts, and set the dilation in IEA to 3. Muon is applied to convolution kernels with a learning rate of $3 \times 10^{-3}$, and AdamW is used for linear projection layers with a learning rate of $5 \times 10^{-4}$. Learning rates are halved at iteration milestones $[200\text{k}, 300\text{k}, 350\text{k}, 375\text{k}, 390\text{k}]$. 

 \section{Comparison of inference time and memory usage} We compare the inference time of our IET model with several state-of-the-art SR methods, including ATD~\cite{zhang2024atd}, IPG~\cite{Tian2024ipg}, and PFT~\cite{long2025pft}, as presented in Tab.~\ref{tab:params_flops_time}. All inference times are measured on a single NVIDIA GeForce RTX 5090 GPU with an output resolution of $256 \times 256$, ensuring a fair and consistent evaluation protocol across methods. At the $\times 3$ and $\times 4$ scales, our model achieves the fastest inference speed among the four approaches, while at the $\times 2$ scale, it is only marginally slower than ATD.
 Moreover, the memory consumption of IET is higher than that of ATD and IPG, but lower than that of PFT.
This is because PFT’s shifted-window mechanism computes attention across two windows, incurring high overhead, especially in early layers, whereas IET assigns each token only one set of individualized neighbors.
This notable efficiency primarily arises from IET’s highly effective attention candidate selection mechanism, which also contributes to its superior reconstruction quality.

\begin{table}[t]
\centering
\small
\caption{Comparison of model size, FLOPs, and inference time across different scales.}
\begin{tabular}{l l c c c c c}
\hline
Scale & Method & Params & FLOPs & Runtime & Memory & PSNR\\
\hline
\multirow{4}{*}{$\times 2$}
& IPG~\cite{Tian2024ipg} & 18.1M & 5.35T & 251ms & 2889M & 34.48/40.24 \\
& ATD~\cite{zhang2024atd} & 20.1M & 6.07T & 143ms & 2565M & 34.70/40.37 \\
& PFT~\cite{long2025pft} & 19.6M & 5.03T & 162ms & 3611M & 34.90/40.49 \\
\rowcolor{gray!15} 
& IET (Ours) & 19.7M & 5.02T & 147ms & 3125M & 35.07/40.61 \\
\hline
\multirow{4}{*}{$\times 3$}
& IPG~\cite{Tian2024ipg} & 18.3M & 2.39T & 151ms & 1669M & 30.36/35.53 \\
& ATD~\cite{zhang2024atd} & 20.3M & 2.69T & 108ms & 1603M & 30.46/35.63 \\
& PFT~\cite{long2025pft} & 19.8M & 2.23T & 114ms & 1802M & 30.56/35.67 \\
\rowcolor{gray!15} 
& IET (Ours) & 19.9M & 2.25T & 106ms & 1788M & 30.81/35.82 \\
\hline
\multirow{4}{*}{$\times 4$}
& IPG~\cite{Tian2024ipg} & 17.0M & 1.30T & 95ms & 875M & 28.13/32.52 \\
& ATD~\cite{zhang2024atd} & 20.3M & 1.52T & 72ms & 802M & 28.17/32.62 \\
& PFT~\cite{long2025pft} & 19.8M & 1.26T & 70ms & 986M & 28.20/32.63 \\
\rowcolor{gray!15} 
& IET (Ours) & 19.8M & 1.26T & 64ms & 970M & 28.43/32.81 \\
\hline
\end{tabular}
\label{tab:params_flops_time}
\end{table}

\section{More Visual Examples}
\begin{figure}[t]
    \centering
    \includegraphics[width=\textwidth]{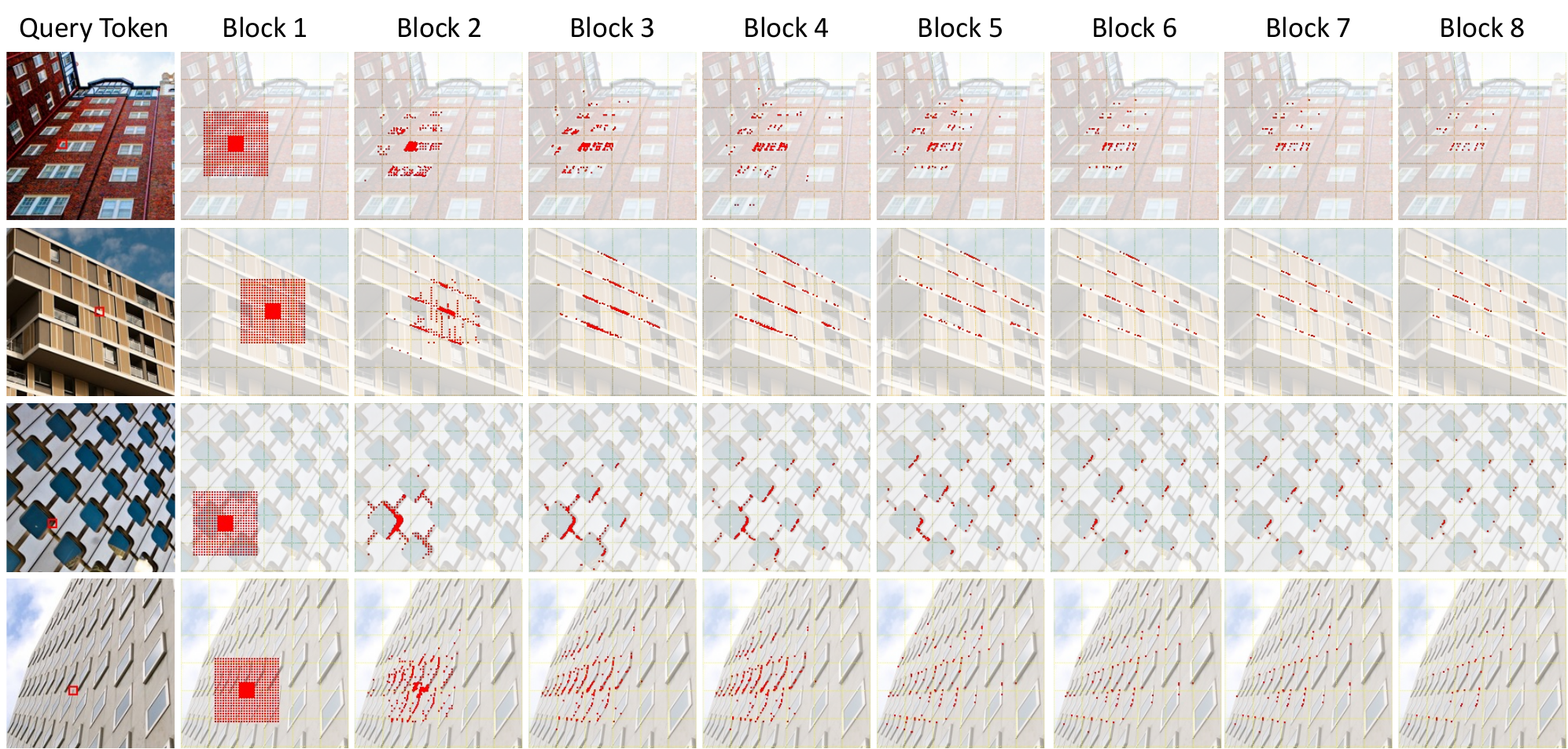}
    \caption{Visualization of the attention candidates initialization and expansion process. The yellow grid represents a 32×32 window, which is the largest window size used among window-based methods.}
    \label{fig:attention_candidates}
\end{figure}

\textbf{Visualization of attention candidates.} The
visualization of the refinement process of attention candidates across different blocks of the IET-light model is shown in Fig.~\ref{fig:attention_candidates}. In the first block, the attention scope is initialized as a dense local and sparse global region centered on each token. As the network deepens, the IEA module gradually expands the attention candidates by aggregate with new two-hop similar tokens while pruning low-similarity neighbors to maintain efficiency with comparable computational cost.

\begin{figure}[t]
    \centering
    \includegraphics[width=\textwidth]{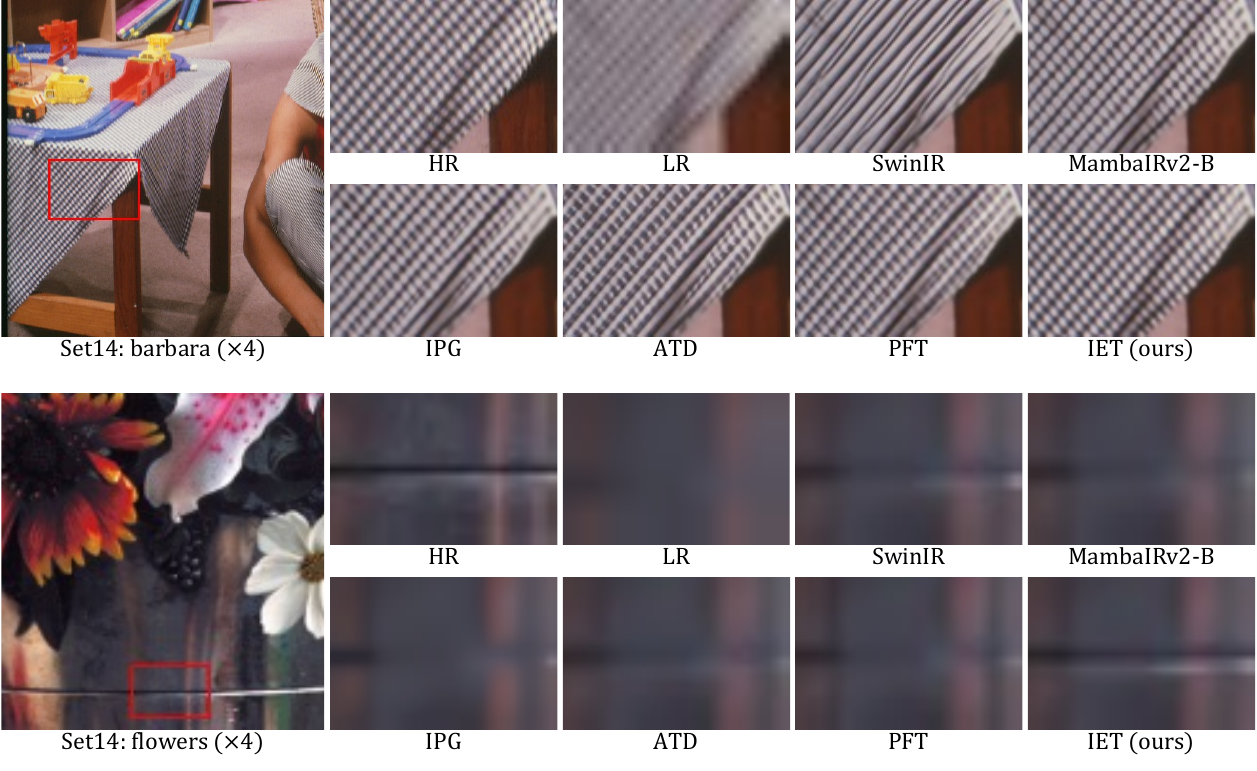}
    \caption{Visual comparisons of \textbf{IET} and other state-of-the-art image super-resolution methods.}
    \label{fig:classical_visualization_1}
\end{figure}

\begin{figure}[h]
    \centering
    \includegraphics[width=\textwidth]{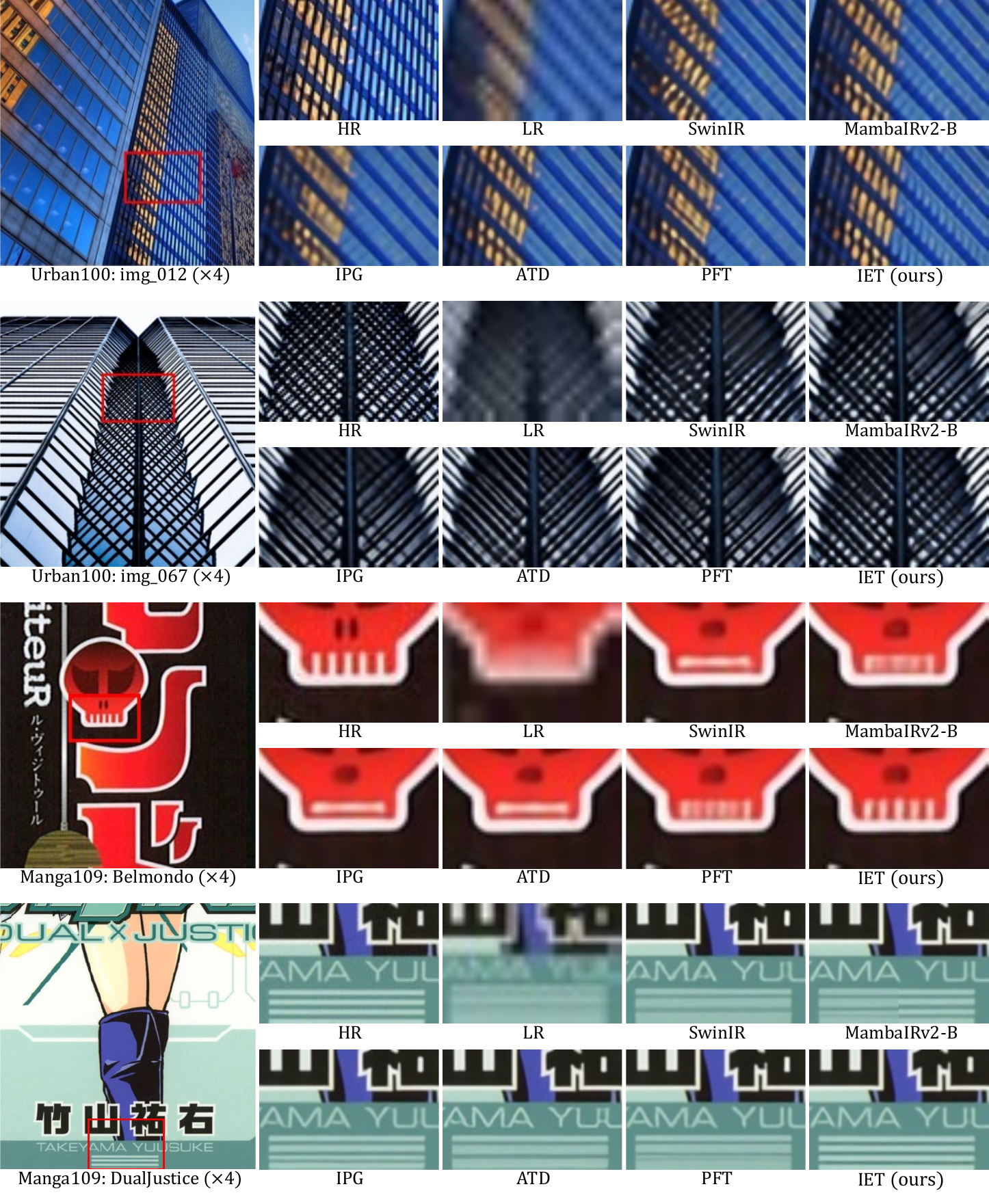}
    \caption{Visual comparisons of \textbf{IET} and other state-of-the-art image super-resolution methods.}
    \label{fig:classical_visualization_2}
\end{figure}

\begin{figure}[h]
    \centering
    \includegraphics[width=\textwidth]{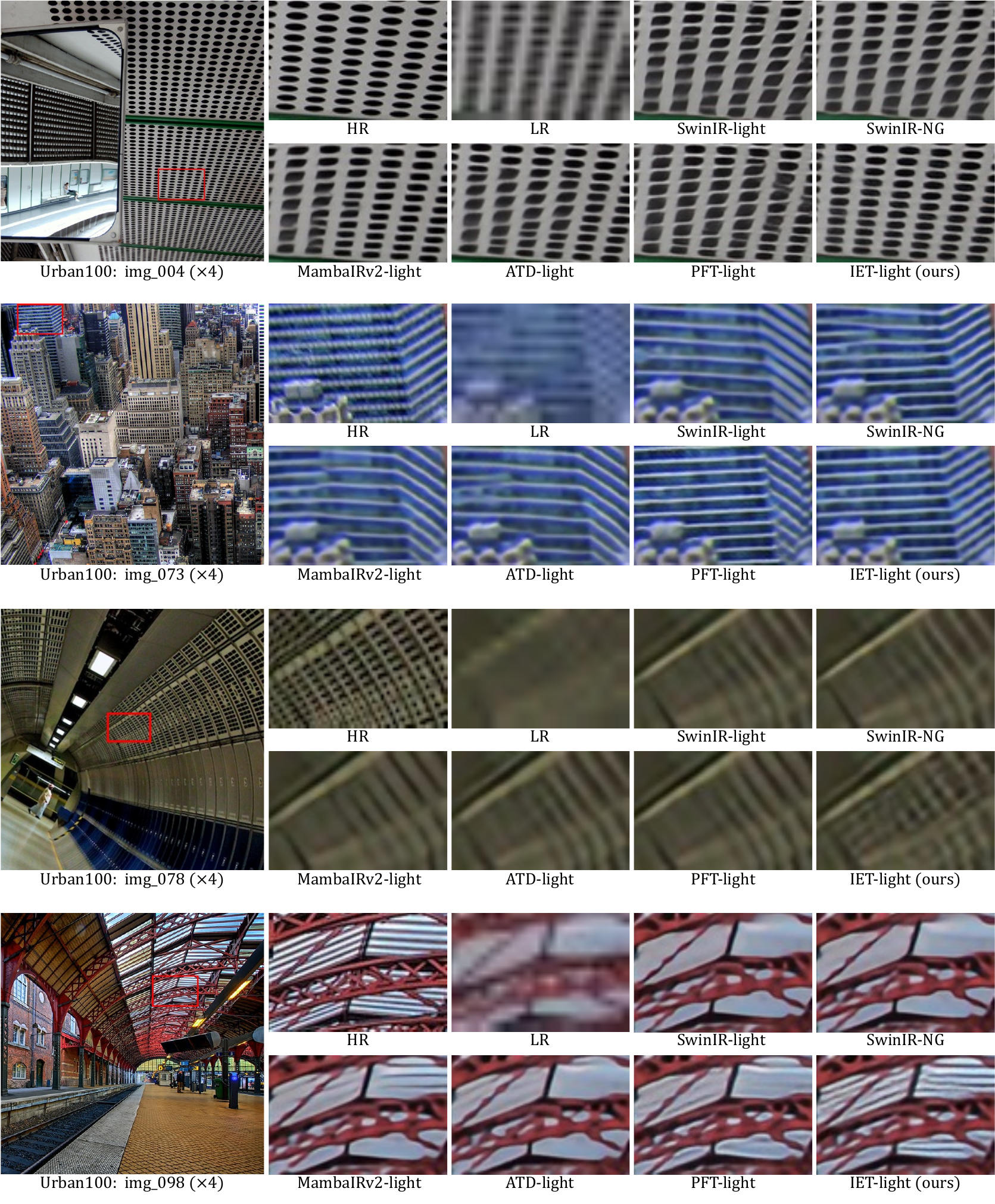}
    \caption{Visual comparisons of \textbf{IET-light} and other state-of-the-art image super-resolution methods.}
    \label{fig:light_visualization_1}
\end{figure}

\begin{figure}[h]
    \centering
    \includegraphics[width=\textwidth]{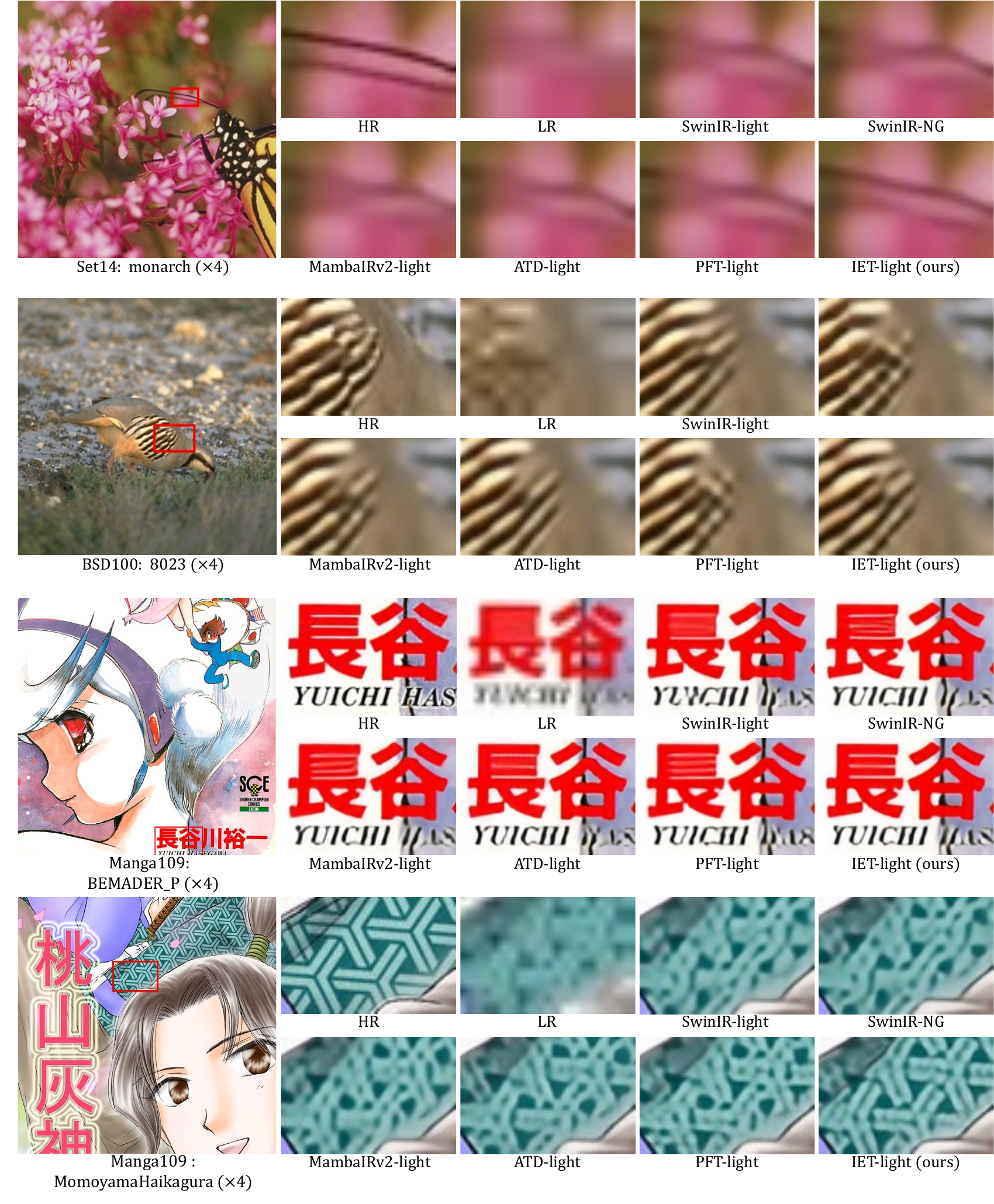}
    \caption{Visual comparisons of \textbf{IET-light} and other state-of-the-art image super-resolution methods.}
    \label{fig:light_visualization_2}
\end{figure}

\smallskip
\noindent\textbf{Visual comparisons of IET.} To qualitatively evaluate the reconstruction performance of our IET and IET-light models in comparison with other methods, we provide visual examples in Fig.~\ref{fig:classical_visualization_1}, Fig.~\ref{fig:classical_visualization_2}, Fig.~\ref{fig:light_visualization_1}, and Fig.~\ref{fig:light_visualization_2}. These comparisons clearly emphasize the strengths of our approach in restoring sharp edges and fine textures from severely degraded low-resolution inputs.

\end{document}